\documentclass{article} % For LaTeX2e
\usepackage{iclr2027_conference,times}

\usepackage{amsmath,amsfonts,bm}

\def\eqref#1{equation~\ref{#1}}
\def\1{\bm{1}}

\DeclareMathAlphabet{\mathsfit}{\encodingdefault}{\sfdefault}{m}{sl}
\SetMathAlphabet{\mathsfit}{bold}{\encodingdefault}{\sfdefault}{bx}{n}

\usepackage{hyperref}
\usepackage{url}
\usepackage{wrapfig}
\usepackage{booktabs,graphicx}
\usepackage{multirow}
\usepackage{algorithm}
\usepackage{algpseudocode}
 \usepackage{booktabs}
\usepackage{multirow}
\usepackage{array}
\usepackage[table]{xcolor}  

\definecolor{avgrow}{gray}{0.93}
\newcommand{\best}[1]{\textbf{#1}}
\newcommand{\snd}[1]{\underline{#1}}

\usepackage{booktabs}
\usepackage{multirow}
\usepackage[table]{xcolor}

\definecolor{overallgray}{gray}{0.95}

\usepackage[normalem]{ulem}

\iclrfinalcopy
\title{Beyond Action Prediction: Learning Search-Free Planning in Latent Space}
\author{%
\begin{minipage}[t]{\dimexpr\textwidth-2\tabcolsep\relax}
    \centering
    \normalfont
    \textbf{Guanchen Du}$^{1}$, \quad
    \textbf{Jinhao Du}$^{2}$, \quad
    \textbf{Yanpeng Sun}$^{3}$, \\[0.2em]
    \textbf{Shan Zhang}$^{4}$, \quad
    \textbf{Na Zhao}$^{3}$, \quad
    \textbf{Zechao Li}$^{1}$ \\[0.6em]
    \small
    $^{1}$Nanjing University of Science and Technology \quad
    $^{2}$University of Oxford \\[0.2em]
    $^{3}$Singapore University of Technology and Design \quad
    $^{4}$Adelaide University\\[0.6em]
    \textbf{Project page:} \url{https://hijp.demohub.live}
\end{minipage}%
}
\begin{document}

\maketitle
\lhead{}
\begin{abstract}
Latent world models learn action-conditioned state transitions in a compact representation space, providing a predictive model for visual planning and control. However, planning with these models typically requires costly test-time optimization over action sequences. We propose to replace this online search with a goal-conditioned inverse model that directly predicts actions from current and goal observations. A key challenge is that action prediction alone neither ensures that goal representations encode action-relevant information nor that predicted actions are consistent with the learned latent prediction model. We introduce Hindsight Inverse JEPA Planning (HIJP), a search-free framework that addresses both aspects during training. HIJP relabels future observations along offline trajectories as goals, providing action supervision across multiple temporal horizons. By retaining gradients through the goal representations, this supervision directly shapes the visual representation. HIJP further feeds predicted actions into the latent predictor model and aligns the resulting next-state representations with encoded successor observations, enforcing action-transition consistency. At inference time, the inverse dynamics model and latent predictor models are applied alternately to generate multi-step action sequences without online search. Across four visual control tasks, HIJP achieves 86.8\% overall success and a 1.9–16.5× planning speedup over search-based planning.

% Latent world models enable visual control, but search-based planning requires substantial test-time computation. Goal-conditioned action prediction offers a search-free alternative; however, existing methods still face two key limitations: first, they do not fully leverage future observations to construct goal supervision across multiple temporal horizons; second, they lack explicit constraints on the state transitions induced by predicted actions, which may therefore be inconsistent with the desired successor states. We propose Hindsight Inverse JEPA Planning (HIJP), which jointly trains a visual encoder, a goal-conditioned inverse dynamics module, and a forward model. HIJP uses hindsight goal relabeling to enrich state–goal supervision across temporal horizons. It further feeds predicted actions into the forward model and aligns the resulting next-state predictions with encoded successor observations, complementing demonstrated-action supervision with an explicit transition constraint. At planning time, the inverse and forward models are applied alternately to synthesize action sequences without online search. Across four visual control tasks and four evaluation configurations, HIJP achieves an overall success rate of 86.8\%, outperforming the strongest baseline \delete{, INTACT,} by 4.9 percentage points. For the same number of replanning calls, HIJP achieves a 1.9–16.5× planning speedup over LeWM.
\end{abstract}

\section{Introduction}
% \begin{wrapfigure}{R}{0.48\columnwidth}
%     \centering
%     \includegraphics[width=\linewidth]{figures/motivation.pdf}
%     \caption {Motivation for HIJP.  (A) With detached goal features, fitting demonstrated actions can still leave a large distance between the predicted next-state latent $\hat z_{t+1}$ and the encoded next observation $z_{t+1}$.
%     (B) HIJP retains gradients through hindsight goals and applies action-transition consistency through $\mathcal L_{\mathrm{cyc}}$, encouraging closer alignment.}
%     \label{fig:motivation}
% \end{wrapfigure}

Latent world models encode high-dimensional visual observations into compact latent representations and learn an action-conditioned latent predictor from offline trajectories, providing a foundation for visual planning and control \citep{zhou2025dinowm,maes2026leworldmodel,zhang2025efficientimagegoalnavigationrepresentative}. 
However, planning with these models typically requires optimizers such as CEM  \citep{rubinstein1999cross} or MPPI \citep{williams2017information} to repeatedly generate and evaluate candidate action sequences at each replanning step, which requires substantial test-time computation \citep {pinneri2021sample}.
Learning a direct mapping from the current state and goal to actions therefore provides a promising alternative for improving the planning efficiency of latent world models \citep{sun2026intactisomorphicintenttoactionlearning, nguyen2026latentgeometrysearchamortizing}.

Recent work has introduced action supervision to improve latent world models' control capabilities. 
SMWM learns action-relevant representations through inverse dynamics supervision \citep{ivashkov2026sensorimotorworldmodelsperception}, while Delta-JEPA \citep{zhang2026deltajepalearningactionsensitiveworld} recovers actions from latent state differences, encouraging representations to retain action-related state changes.
GC-IDM \citep{nguyen2026latentgeometrysearchamortizing} further learns a goal-conditioned action mapping on frozen representations, while INTACT  \citep{sun2026intactisomorphicintenttoactionlearning}incorporates local and future-goal action supervision into end-to-end training, supporting search-free recursive control. 
% However, these training approaches still have limitations in constructing goal supervision and jointly constraining action and state prediction.
These approaches demonstrate the value of action supervision for goal-conditioned control. However, accurate action prediction alone does not necessarily provide all the supervision needed for search-free planning.

% motivation. too long, will be refined ok

We identify two aspects that remain underconstrained. 
First, goal-conditioned action supervision should not only train an inverse model to predict actions, but also help shape visual representations to capture how actions relate current states to future goals.
Offline trajectories naturally contain such relations at multiple temporal scales: for the same current state and demonstrated action, different future observations define goals at different temporal horizons.
However, existing training schemes  \citep{sun2026intactisomorphicintenttoactionlearning,wang2026decision}  exploit only a subset of these relations, for example by conditioning action prediction on a fixed future endpoint.
More generally, treating the goal representation only as a conditioning target limits the extent to which action supervision can shape goal-side visual features.
Second, accurate action prediction alone does not ensure that the predicted action induces the correct transition under the learned latent predictor.
The inverse dynamics model and the latent predictor model are typically supervised separately, with the latter conditioned on demonstrated actions  \citep{ivashkov2026sensorimotorworldmodelsperception,sun2026intactisomorphicintenttoactionlearning}.
During search-free rollout planning, however, these two mappings are applied sequentially: the inverse dynamics model predicts an action toward the goal, and the latent predictor model uses this predicted action to advance the latent state for subsequent action generation.
The resulting inverse-to-predictor composition is therefore directly involved in planning, yet is not explicitly constrained by separate action and latent prediction objectives.
This motivates extending action supervision beyond action prediction itself, using it both to shape goal-conditioned representations and to couple predicted actions with their induced latent transitions.

% First, methods that use the terminal state of a window as a fixed goal\citep{sun2026intactisomorphicintenttoactionlearning,wang2026decision} do not fully utilize other future states \blue{within the window} to construct goal-conditioned action supervision, limiting the state–goal combinations provided by each sampled window. Second, methods that separately supervise action prediction and state transitions conditioned on demonstrated actions \citep{ivashkov2026sensorimotorworldmodelsperception,sun2026intactisomorphicintenttoactionlearning} constrain action prediction errors and forward prediction errors, but do not explicitly constrain the next latent state produced by passing the predicted action through the forward model.
% In inverse dynamics-based rollout planning, this predicted state is subsequently used to generate further actions, and its error may affect subsequent decisions. Therefore, this paper focuses on two questions: how to construct richer state–goal combinations within finite trajectory windows to supervise current-action learning for goals at different temporal horizons; and how to explicitly constrain the forward transition outcomes of predicted actions to provide action generation with direct feedback at the state-prediction level.

% our method.
Based on this principle, we propose Hindsight Inverse JEPA Planning (HIJP), which jointly trains a visual encoder, a goal-conditioned inverse dynamics model, and a latent predictor model end-to-end. HIJP relabels future observations within each trajectory window as goals, exposing the inverse objective to diverse state--goal relations across temporal horizons. Gradients are retained through the goal representations, allowing inverse-action supervision to directly shape the visual features used for goal-conditioned action prediction. We further feed the predicted actions into the latent predictor and align the resulting successor latents with the encodings of the next observations. This action-transition consistency objective jointly updates the inverse and predictor models, directly coupling action generation with its induced latent transition. In this way, HIJP trains the same inverse-to-predictor mapping that is repeatedly applied during planning. At test time, the inverse dynamics module and latent predictor model alternate to generate multi-step action sequences, which are executed through closed-loop replanning without online action search.

% To answer the questions, we propose HIJP (Hindsight Inverse JEPA Planning), which jointly trains an encoder, a goal-conditioned inverse dynamics module, and a forward model end-to-end. To enrich the goal-conditioned supervision provided by each trajectory window, we adopt hindsight goal relabeling\citep{andrychowicz2017hindsight,ghosh2019learning}, sampling starting points and their future observations within each window. We use the future observations as goals and the demonstrated actions at the starting points to supervise the inverse module, thereby expanding state–goal combinations and supporting current-action learning across different temporal horizons. We further feed the actions generated under these different future-goal conditions into the forward model and align the predicted next latent states with the encodings of the next observations in the trajectories, so that each action prediction is constrained by both demonstrated-action supervision and successor-state prediction errors. At test time, the inverse dynamics module and the forward model alternate to generate multi-step action sequences, which are executed through closed-loop replanning without online action search.

We evaluate HIJP on four visual control tasks under four configurations of goal temporal offsets and execution budgets. HIJP achieves an overall success rate of \(86.8\%\), outperforming the strongest baseline by \(4.9\) percentage points. Ablations on hindsight goal sampling, training objectives, and gradient routing examine how goal-side representation learning and inverse - predictor coupling contribute to control performance. Beyond success rates, we also evaluate efficiency and the temporal smoothness of executed actions. With matched numbers of replanning calls, HIJP achieves a \(1.9\)–\(16.5\times\) planning speedup over LeWM. Under the most challenging setting, its executed actions exhibit lower jerk than those produced by CEM using the same representations, indicating smoother executed controls.

\section{Related work}
\label{section:related}
\paragraph{Latent World Models}
Latent world models learn action-conditioned changes in representation space to support goal-directed planning from offline visual trajectories. Existing approaches span pretrained visual features, end-to-end representation learning, and large-scale video pretraining \citep{zhou2025dinowm,maes2026leworldmodel,vjepa2Mido2025, bo2026diagram}.
Systematic studies further investigate how architectures, training objectives, and planning algorithms affect control performance \citep{terver2025drives}. \citet{gao2026fast} reduce forward-rollout costs through parallel prediction conditioned on action prefixes. \citet{wang2026decision} use inverse dynamics and goal-conditioned action supervision to improve planning representations, while retaining CEM for control.
HIJP builds on latent forward dynamics and focuses on learning a goal-conditioned action mapping that can be directly deployed for control.

\paragraph{Inverse Dynamics and Goal-Conditioned Action Learning}
Inverse dynamics and action-supervised objectives encourage representations to retain action-relevant state changes.
These objectives can also be combined with physical-state alignment to improve representations for control \citep{ivashkov2026sensorimotorworldmodelsperception,zhang2026deltajepalearningactionsensitiveworld, liu2026physicallygroundedjepaworld}.
Hindsight learning relabels achieved states along trajectories as goals, providing supervision for goal-conditioned policies and inverse dynamics learning \citep{andrychowicz2017hindsight,ghosh2019learning,sun2019policy}.
Goal-conditioned action-sequence prediction has also been used for visual planning \citep{paster2020planning}. Related theoretical studies analyze the benefits of inverse dynamics through bias–variance trade-offs and learning-problem complexity \citep{schafer2026does,morin2026sample}.
Within latent world models, GC-IDM\citep{nguyen2026latentgeometrysearchamortizing} learns a goal- and horizon-conditioned action mapping on frozen representations, while INTACT  \citep{sun2026intactisomorphicintenttoactionlearning} jointly learns local and goal-conditioned action prediction to support search-free recursive control. HIJP follows this direction by enriching supervision through future-goal relabeling within trajectory windows.

\paragraph{Learned and Amortized Planning}
Planning computation can be shifted to training by learning action priors, latent subgoals, or plan-update rules \citep{wang2026prism,cheng2026sage,masip2026ff,sommer2026reinforced, tang2025artemis}.
Another line of work first predicts future states or generates state trajectories and then recovers actions through inverse dynamics \citep{tian2025predictive,xie2025latent,zeng2026dtagtdirectiontopologyawaregraph,ajay2022conditional,huang2026leflow}. \citet{rakhimov2026qantara} jointly train state and action flows, allowing a single model to support multiple control inference modes.
HIJP alternates goal-conditioned action prediction with forward latent-state prediction to construct multi-step plans without online action search.

\paragraph{Consistency and Goal-Aware Planning Objectives}
Consistency objectives also connect model prediction with planning. 
\citet{seo2026acidactionconsistencyinverse} incorporate action-recovery errors from predicted transitions into the online planning cost, while \citet{bai2026temporal} combine temporal-progress learning with multi-step rollout consistency supervision.
HIJP focuses on the composed output of the goal-conditioned action module and the forward model. During training, it aligns the successor latent state predicted under the generated action with the encoding of the next observation in the recorded trajectory, allowing state-prediction errors to supervise action learning directly.

\section{Methodology}
\label{section:method}
\begin{figure}[t]
    \centering
    \includegraphics[width=\textwidth]{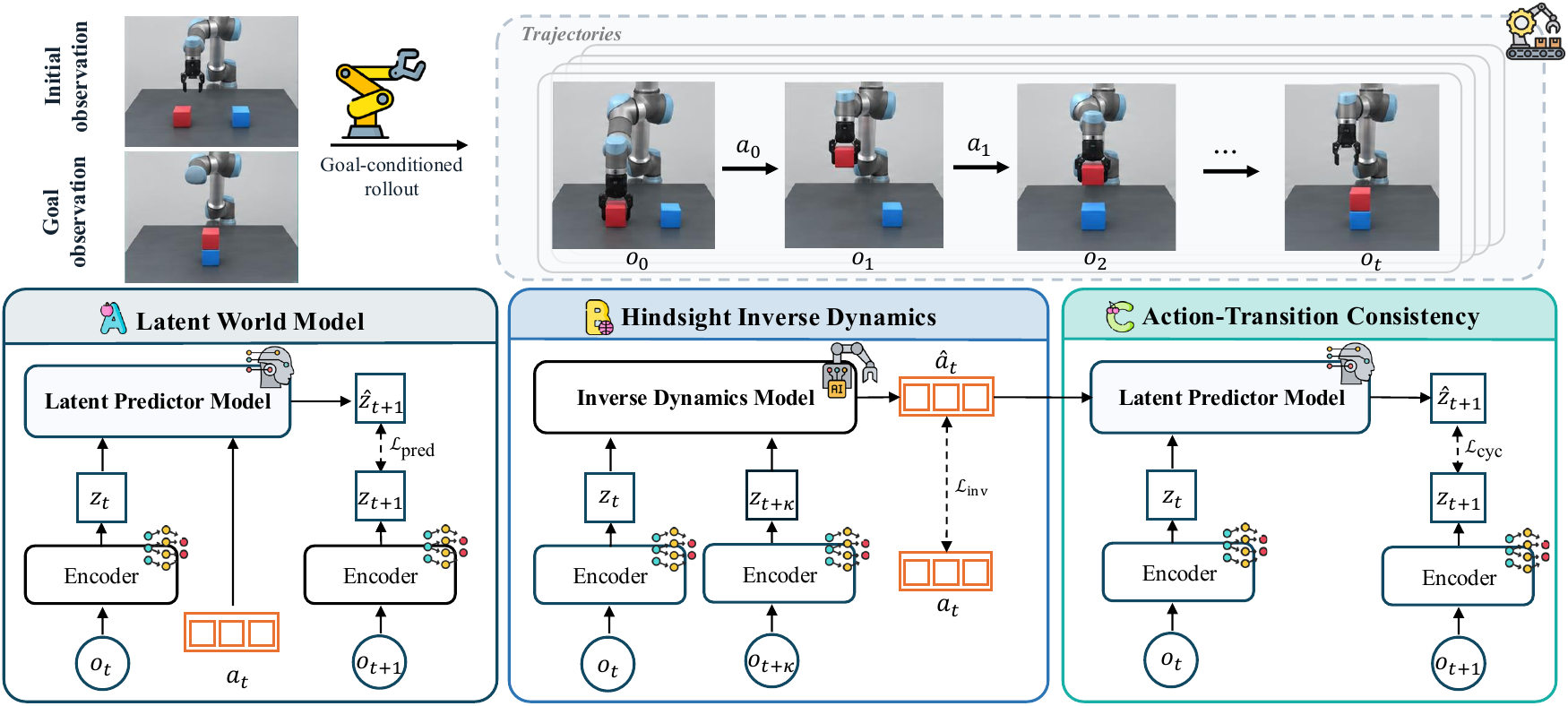}
    \caption{Overview of the HIJP training framework. (A) The latent world model predicts the next latent state from the current state and action. (B) The hindsight-conditioned inverse dynamics model infers the action from current and future latent states. (C) The inferred action is reused to enforce action–transition consistency. All modules are jointly optimized using the prediction, regularization, inverse-dynamics, and action transition consistency losses.}
    \label{fig:overview}
\end{figure}

Our goal is to extend action supervision beyond action prediction itself, such that it also shapes goal-conditioned visual representations and constrains the latent transitions induced by predicted actions.
HIJP jointly learns a visual encoder $f_\theta$, a latent predictor model $g_\phi$, and a goal-conditioned inverse dynamics model $h_\xi$. 
Given offline visual trajectories, the inverse dynamics model learns to predict actions toward future goals, while the predictor predicts action-conditioned latent transitions. 
We couple these components during training to supervise the local inverse-to-predictor composition repeatedly used in search-free planning.
Figure~\ref{fig:overview} provides an overview.
% Our method belongs to the family of latent model-based control methods. Unlike prior approaches \citep{zhou2025dinowm} that perform online action search with sampling-based optimizers, HIJP learns a hindsight-conditioned inverse dynamics module for direct goal-conditioned action prediction. Meanwhile, the predicted actions are passed through the forward model, and the resulting successor latents are aligned with encoded successor observations. 

% Figure~\ref{fig:overview} illustrates the architecture and training objectives. HIJP comprises a visual encoder $f_\theta$, a forward predictor $g_\phi$, and an inverse dynamics module $h_\xi$. We next describe the model, training objective, and planning procedure.

\subsection{Problem Formulation}
Let $\mathcal{D}=\{\tau^{(i)}\}_{i=1}^{N}$ be a dataset of trajectories, where 
\begin{equation}
    \tau = (o_0, \mathbf{a}_0, o_1, \ldots, \mathbf{a}_{T-1}, o_T). 
\end{equation}
Here $o_t$ is a visual observation and $\mathbf{a}_t$ is the action taken at time $t$.
Given a current observation $o_t$ and a goal observation \(o_g\), the task is to produce actions that reach the goal within a fixed interaction budget. Training uses recorded observations and actions, with future observations in each trajectory serving as hindsight goals, where reward annotations are not required.
Our planning procedure repeatedly composes two mappings. An inverse model predicts an action from the current latent state and goal latent state, and a latent predictor model predicts the latent state induced by that action. This composition motivates training the models together: action prediction must identify a useful action, while the resulting transition must agree with the observed dynamics.

\subsection{Latent World Model}
Following the prior work \citep{maes2026leworldmodel}, a visual encoder \(f_\theta:\mathcal O\rightarrow\mathbb R^d\) maps observations to latent states \(z_t=f_\theta(o_t)\), and an action-conditioned latent predictor model \(g_\phi\) predicts the next latent state from a causal context of latent states and actions:
\begin{equation}
\hat z_{t+1} = g_\phi\!\left(z_{t-m+1:t},a_{t-m+1:t}\right),
\qquad
z_{t+1}=f_\theta(o_{t+1}).
\end{equation}
We train the latent prediction model on demonstrated transitions using
\begin{equation}
\mathcal L_{\mathrm{pred}} = \operatorname{MSE}(\hat z_{t+1},z_{t+1}).
\end{equation}
We also apply \(\mathcal L_{\mathrm{sigreg}}\) to the encoded observations in each training window to discourage representational collapse. 
We use a ViT-Tiny encoder trained from scratch and a causal Transformer model with context length \(m=3\). An action encoder supplies the latent prediction model with action embeddings. Architecture and optimization details are given in Appendix ~\ref{app:impl}. The same model also accepts the shorter context used for the one-step consistency objective below.

The latent prediction model learns how demonstrated actions transform latent states. Our next objective is complementary: to learn how actions can be inferred from current and goal representations, while allowing this action supervision to shape the representations themselves.

\subsection{Hindsight-conditioned Inverse Dynamics}
A trajectory contains multiple valid future goals for a given observation. To expose these relations during training, we sample a window of \(W\) observations, an anchor \(i\) and a future offset \(\kappa\):
\begin{equation}
i\sim\operatorname{Unif}\{0,\ldots,W-2\},
\qquad
\kappa\mid i\sim\operatorname{Unif}\{1,\ldots,W-1-i\}.
\end{equation}
Each sampled pair therefore defines a goal-conditioned action-learning problem at a potentially different temporal horizon. Unlike using only the endpoint of a trajectory window, this construction exposes the inverse objective to diverse relations between current observations and future goals using the same offline data.
The observation at \(i+\kappa\) becomes the hindsight goal. The goal-conditioned inverse model predicts the demonstrated action at the anchor:
\begin{equation}
\hat a_i=h_\xi(z_i,z_{i+\kappa}),\qquad
\mathcal L_{\mathrm{inv}} =\operatorname{MAE}(\hat a_i,a_i).
\end{equation}
Although the sampled goal may be several steps ahead, the target is the immediate action \(a_i\), not the intervening action sequence. Each training window therefore supplies action supervision under goals at different temporal offsets.

Gradients from \(\mathcal L_{\mathrm{inv}}\) flow through both \(z_i\) and \(z_{i+\kappa}\). Consequently, the inverse objective trains not only \(h_\xi\), but also the visual features used to represent the current observation and the future goal. We implement \(h_\xi\) as an MLP operating on the concatenated latent pair.

% 总起逆动力学
% To directly predict the immediate action that moves the agent toward a given goal, we augment the world model with an inverse-dynamics module, instantiated as a feed-forward, goal-conditioned inverse-dynamics head:
% \begin{equation}
%     h_\xi:\ \mathbb{R}^{d}\times\mathbb{R}^{d}\rightarrow\mathbb{R}^{d_a},
%     \quad \mathbf{a}_i \in \mathbb{R}^{d_a},
%     \qquad \hat{\mathbf{a}}_i=h_\xi(z_i,z_{i+\kappa}),
% \end{equation}
% which maps a current latent and a goal latent to the demonstrated immediate action on a trajectory segment that reaches the future observation.

%具体介绍逆动力学模块的做法
% We use a hindsight relabeling strategy. Given a fixed window of $W$ consecutive steps, we sample an anchor $i \sim \operatorname{Unif} \{0, \ldots, W-2 \}$. Conditioned on $i$, we then sample a hindsight-goal offset $\kappa \sim \operatorname{Unif}\{1, \ldots, W-1-i\}$, yielding the goal latent $z_{i+\kappa}$. The inverse-dynamics head is trained to predict the action $\mathbf{a}_i$ from the pair $(z_i, z_{i+\kappa})$.

% We implement $h_\xi$ as a $3$-layer MLP with a hidden width of $512$ and GELU activations, taking the concatenated latent pair ($2d$ dimensions) as input and regressing the action.

% 新加的action transition consistency

\subsection{Action Transition Consistency}
The inverse model can predict an action close to the demonstrated action without ensuring that this prediction produces the observed transition under the learned predictor model. This distinction matters at planning time, where the model receives predicted rather than demonstrated actions. We therefore pass the predicted action through \(g_\phi\):
\begin{equation}
\tilde z_{i+1}
=
g_\phi\!\left(z_i,\hat a_i\right),
\qquad
\hat a_i=h_\xi(z_i,z_{i+\kappa}),
\end{equation}
and align the result with the encoded one-step successor:
\begin{equation}
\mathcal L_{\mathrm{cyc}}
=
\operatorname{MSE}
\!\left(\tilde z_{i+1},\operatorname{sg}[z_{i+1}]\right),
\end{equation}
where \(\operatorname{sg}\) denotes stop-gradient. 
This objective uses the one-step successor \(z_{i+1}\) even when the hindsight goal \(z_{i+\kappa}\) is farther away.
It allows \(L_{\mathrm{cyc}}\) to update the inverse head, action encoder, latent prediction model, and the anchor-side visual representation. Thus, the consistency objective does not merely use the prediction model to provide an error signal to the inverse dynamics model; it jointly adapts action generation and the latent predictor around their composed prediction.
This directly supervises the inverse-to-latent predictor transition that is repeatedly composed during search-free planning.
% The inverse dynamics supervises the action predicted from the current latent state and a hindsight goal. We additionally use the forward predictor to check the transition induced by that action. Given $\hat{\mathbf a}_i=h_\xi(z_i,z_{i+\kappa})$, we define
% \begin{equation}
%     \tilde z_{i+1}
%     = g_\phi(z_i,\hat{\mathbf a}_i),
%     \qquad
%     \hat{\mathbf a}_i=h_\xi(z_i,z_{i+\kappa}).
% \end{equation}

% Although the hindsight goal $z_{i+\kappa}$ may be several steps ahead, the predicted successor $\tilde z_{i+1}$ is supervised by the encoded one-step successor $z_{i+1}$. 
% This couples goal-conditioned action prediction to the learned forward dynamics.

\subsection{Joint Objective and Search-free Planning}
\textbf{Joint training.}
The full training objective combines latent prediction, representation regularization, hindsight-conditioned action prediction, and action-transition consistency:
% The model is trained end-to-end by minimizing a weighted sum of four loss terms,
\begin{equation}
    \mathcal{L}=\mathcal{L}_{\text{pred}}
    +\lambda_{\text{reg}}\,\mathcal{L}_{\text{sigreg}}
    +\lambda_{\text{inv}}\,\mathcal{L}_{\text{inv}}
    +\lambda_{\text{cyc}}\,\mathcal{L}_{\text{cyc}} .
    \label{eq:total}
\end{equation}
All components are optimized jointly, and all terms are averaged over their corresponding batch, temporal, sampled-pair, and feature dimensions before weighting.
For all tasks, the loss weights are set to $\lambda_{\mathrm{reg}}=0.09$, $\lambda_{\mathrm{inv}}=1.0$, and
$\lambda_{\mathrm{cyc}}=0.05$ for TwoRoom and $0.10$ for the remaining tasks.

\textbf{Search-free planning.}
At planning time, we use the inverse dynamics model and latent prediction models in the same order in which they are coupled during training. At each replanning point, the latest observation and fixed goal observation are encoded as $z_t=f_\theta(o_t)$ and $z_g=f_\theta(o_g)$. Starting from $\hat z_0=z_t$, we construct an \(H\)-step plan by alternating goal-conditioned action prediction and latent prediction:
\begin{equation}
a_j=h_\xi(\hat z_j,z_g),\qquad
\hat z_{j+1}
=
g_\phi\!\left(\hat z_{\ell_j:j},a_{\ell_j:j}\right),
\quad
\ell_j=\max(0,j-m+1),
\quad j=0,\ldots,H-1.
\end{equation}
The inverse model therefore predicts each action from the current imagined state and fixed goal, while the latent prediction model advances the imagined state under that predicted action. We use \(H=5\), execute the resulting action sequence, observe the environment again, and repeat the procedure in a receding-horizon manner. Each replanning call requires only one inverse-to-latent prediction rollout, without sampling or iteratively refining candidate action sequences.

\section{Experiments}
\label{section:exp}
We conduct experiments on four tasks to answer the following key questions: (i) whether our method achieves strong goal-reaching performance, particularly for long-horizon goals; (ii) how much inference speed is gained by replacing the CEM search at test time with a single amortized rollout; (iii) how hindsight goal relabeling, inverse-action supervision, gradient routing, and predicted-action transition consistency contribute to control performance; and (iv) how HIJP and CEM differ in
latent-space goal progress and executed-action smoothness.

\subsection{Experiment setup}
\label{sec:exp-setup}

\paragraph{Benchmarks and Tasks}
\label{sec:exp-benchmarks}
We evaluate on four goal-conditioned control benchmarks with distinct dynamics: \emph{TwoRoom} ($2$D navigation), \emph{Reacher} (DMControl reacher), \emph{PushT} (planar pushing), and \emph{OGCube} (OGBench cube manipulation).
All tasks use $224{\times}224$ RGB observations, continuous actions, and observation-specified goals. The dataset information is provided in Appendix~\ref{app:datasets}.
We use the offline datasets released by LeWM~\citep{maes2026leworldmodel} without modification.
To vary difficulty, we set each goal to the environment state $K\in\{10,25,50,75\}$ steps ahead of the start and assign a corresponding interaction budget $B\in\{25,50,100,100\}$, defining the \emph{easy}, \emph{default} (LeWM's\citep{maes2026leworldmodel} original goal-setting horizon and interaction budget), \emph{hard}, and \emph{extreme} tiers, respectively.
All methods are evaluated in the original simulator on the same start--goal instances, with $50$ episodes per task--tier combination. An episode counts as successful only when the simulator's native success condition is met; we use no latent-space distance threshold. We report success rate as the primary metric and planning time for the efficiency study, with additional details in Appendix~\ref{app:eval-protocol}.

\paragraph{Baselines}
\label{sec:exp-baselines}

We compare against eight baselines: Random, GCBC~\citep{ghosh2021learning}, DINO-WM~\citep{zhou2025dinowm}, PLDM~\citep{sobal2025stresstesting}, LeWM~\citep{maes2026leworldmodel}, SMWM~\citep{ivashkov2026sensorimotorworldmodelsperception}, delta-JEPA~\citep{zhang2026deltajepalearningactionsensitiveworld} and INTACT\citep{sun2026intactisomorphicintenttoactionlearning}.
Random provides a chance-level reference; GCBC and INTACT provide direct goal-conditioned controllers, with INTACT representing the closest end-to-end search-free baseline. The remaining methods are latent world-model planners that rely on CEM- or MPPI-based search at test time.
All learned methods are trained on the same offline datasets and evaluated under the same protocol.
All CEM-based comparisons use the shared search and execution
protocol specified in Appendix~\ref{app:cem-protocol}.

\paragraph{Implementation Details}
\label{sec:exp-impl}
HIJP combines a latent world model with an amortized inverse head trained end-to-end.
At planning time, it produces an $H=5$ step latent plan with a single amortized rollout, executes the corresponding actions in a receding-horizon manner, and then replans from the latest observation. Unlike the CEM-based world-model baselines, HIJP uses no test-time search. Architecture and optimization details are provided in Appendix~\ref{app:impl}.

\subsection{Planning Performance}

% We first assess whether HIJP consistently achieves goals more reliably than current methods. 
We first examine how HIJP's goal-reaching performance compares with the baselines.
As detailed in Section \ref{sec:exp-setup}, we test HIJP and all baseline methods on each task-tier combination, using 50 episodes per pair, and report success rates in Table~
\ref{tab:main}. 

\begin{table*}[t]
\centering
\caption{Success rates (\%) across four tasks and four difficulty tiers. Avg. denotes the arithmetic mean across the four difficulty tiers; Overall Avg. averages all 16 task--tier combinations with equal weight.
The best and second-best results in each row are shown in \textbf{bold} and \underline{underlined text}.}
\label{tab:main}

% 使用模板默认字体，字号比正文小一级
\footnotesize
\fontsize{7.75pt}{8.225pt}\selectfont
\normalfont

% 基础间距较小，剩余宽度由 \extracolsep{\fill} 均匀分配
\setlength{\tabcolsep}{2pt}
\renewcommand{\arraystretch}{1.05}

\begin{tabular*}{\textwidth}{
    @{\extracolsep{\fill}}
    ll
    *{9}{c}
    @{}
}
\toprule
& & Reference
& \multicolumn{7}{c}{Baselines}
& Ours \\
\cmidrule(lr){3-3}
\cmidrule(lr){4-10}
\cmidrule(lr){11-11}

Task & Tier
& Random
& GCBC
& PLDM
& DINO-WM
& LeWM
& SMWM
& $\Delta$-JEPA
& INTACT
& \textbf{HIJP} \\
\midrule

% ======================== TwoRoom ========================

\multirow{5}{*}{TwoRoom}
& Easy
& 66\%
& \best{100\%}
& \best{100\%}
& \best{100\%}
& \snd{96\%}
& \snd{96\%}
& 76\%
& \best{100\%}
& \best{100\%} \\

& Default
& 24\%
& \best{100\%}
& 88\%
& \best{100\%}
& 88\%
& \snd{96\%}
& 38\%
& \best{100\%}
& \best{100\%} \\

& Hard
& 8\%
& \best{100\%}
& 90\%
& \snd{98\%}
& 62\%
& 86\%
& 48\%
& 96\%
& \best{100\%} \\

& Extreme
& 2\%
& \best{100\%}
& 96\%
& 74\%
& 50\%
& 72\%
& 24\%
& 90\%
& \snd{98\%} \\

\cmidrule(l){2-11}
& \textbf{Avg.}
& 25.0\%
& \best{100.0\%}
& 93.5\%
& 93.0\%
& 74.0\%
& 87.5\%
& 46.5\%
& 96.5\%
& \snd{99.5\%} \\

\midrule

% ======================== Reacher ========================

\multirow{5}{*}{Reacher}
& Easy
& 12\%
& 4\%
& \best{70\%}
& 46\%
& 40\%
& 34\%
& 14\%
& 44\%
& \snd{48\%} \\

& Default
& 10\%
& 0\%
& 72\%
& 78\%
& 76\%
& 46\%
& 20\%
& \best{100\%}
& \snd{92\%} \\

& Hard
& 6\%
& 0\%
& 72\%
& \snd{98\%}
& 96\%
& 62\%
& 52\%
& \best{100\%}
& 92\% \\

& Extreme
& 6\%
& 0\%
& 84\%
& \snd{94\%}
& 84\%
& 58\%
& 40\%
& \best{96\%}
& \snd{94\%} \\

\cmidrule(l){2-11}
& \textbf{Avg.}
& 8.5\%
& 1.0\%
& 74.5\%
& 79.0\%
& 74.0\%
& 50.0\%
& 31.5\%
& \best{85.0\%}
& \snd{81.5\%} \\

\midrule

% ======================== PushT ========================

\multirow{5}{*}{PushT}
& Easy
& 14\%
& \snd{94\%}
& \snd{94\%}
& \best{98\%}
& 78\%
& 80\%
& 72\%
& \best{98\%}
& 90\% \\

& Default
& 2\%
& 74\%
& 42\%
& 74\%
& \best{88\%}
& 56\%
& 64\%
& \snd{80\%}
& \snd{80\%} \\

& Hard
& 2\%
& 40\%
& 22\%
& 56\%
& \snd{58\%}
& 30\%
& 36\%
& 54\%
& \best{66\%} \\

& Extreme
& 0\%
& 4\%
& 0\%
& \snd{28\%}
& 12\%
& 16\%
& 0\%
& 6\%
& \best{42\%} \\

\cmidrule(l){2-11}
& \textbf{Avg.}
& 4.5\%
& 53.0\%
& 39.5\%
& \snd{64.0\%}
& 59.0\%
& 45.5\%
& 43.0\%
& 59.5\%
& \best{69.5\%} \\

\midrule

% ======================== OGCube ========================

\multirow{5}{*}{OGCube}
& Easy
& 68\%
& \snd{98\%}
& 74\%
& 92\%
& \snd{98\%}
& \snd{98\%}
& 88\%
& \best{100\%}
& \snd{98\%} \\

& Default
& 44\%
& 90\%
& 56\%
& 78\%
& 82\%
& 88\%
& 74\%
& \best{100\%}
& \snd{96\%} \\

& Hard
& 26\%
& 58\%
& 28\%
& 62\%
& 56\%
& \snd{70\%}
& 54\%
& \snd{70\%}
& \best{96\%} \\

& Extreme
& 40\%
& 76\%
& 50\%
& 72\%
& 70\%
& \snd{86\%}
& 74\%
& 76\%
& \best{96\%} \\

\cmidrule(l){2-11}
& \textbf{Avg.}
& 44.5\%
& 80.5\%
& 52.0\%
& 76.0\%
& 76.5\%
& 85.5\%
& 72.5\%
& \snd{86.5\%}
& \best{96.5\%} \\

% ======================== Overall ========================

\specialrule{0.08em}{0.4em}{0.3em}
\rowcolor{overallgray}
\multicolumn{2}{c}{\textbf{Overall Avg.}}
& 20.6\%
& 58.6\%
& 64.9\%
& 78.0\%
& 70.9\%
& 67.1\%
& 48.4\%
& \snd{81.9\%}
& \best{86.8\%} \\

\bottomrule
\end{tabular*}
\end{table*}

Table~\ref{tab:main} shows that HIJP achieves the highest overall success rate of 86.8\%, exceeding the strongest baseline by 4.9 percentage points.
The gains are most pronounced in challenging manipulation settings.
On PushT, HIJP achieves 66\% and 42\% success in the hard and extreme tiers, respectively, outperforming the best baseline in each tier by 8 and 14 percentage points.
On OGCube, it maintains 96\% success in both tiers, with corresponding gains of 26 and 10 percentage points.
On TwoRoom and Reacher, HIJP remains competitive, ranking second in average success on both tasks.
The stronger gains on manipulation are consistent with our design to extend action supervision beyond action prediction itself, shaping goal-conditioned visual representations and constraining the latent transitions induced by predicted actions.
We examine the contributions of these two design choices through the ablations in Section~\ref{sec:exp-ablation}.

\subsection{Planning Efficiency}
\label{sec:exp-speed}
% 介绍实验的重要性 以及实验是怎么做的
To assess planning efficiency, we compare HIJP with LeWM, a CEM-based world-model planner, under matched numbers of replanning calls. Following Section~\ref{sec:exp-setup}, we report total wall-clock planning time for each task–tier combination, with all episodes processed as a single vectorized batch on an RTX~4090 (Table~\ref{tab:speed}).

\begin{table}[t]
\centering
\caption{
Total wall-clock planning time (seconds) for LeWM and HIJP.
Speedup denotes LeWM's planning time divided by HIJP's.  Avg. reports the arithmetic mean across difficulty tiers, including the per-tier speedup ratios. Lower planning times are shown in \textbf{bold}.
}
\label{tab:speed}

\footnotesize
% \normalfont
\fontsize{7.75pt}{8.225pt}\selectfont
\setlength{\tabcolsep}{2pt}
\renewcommand{\arraystretch}{1.03}

\begin{tabular*}{0.75\textwidth}{
    @{\extracolsep{\fill}}
    ll
    *{9}{c}
    @{}
}
\toprule
\multirow{2}{*}[-1pt]{Task}
& \multirow{2}{*}[-1pt]{Method}
& \multicolumn{4}{c}{Difficulty tier}
& \multirow{2}{*}[-1pt]{Avg.} \\
\cmidrule(lr){3-6}
& & Easy & Default & Hard & Extreme & \\
\midrule

% ======================== TwoRoom ========================

\multirow{3}{*}{TwoRoom}
& LeWM
& 69.8
& 124.9
& 254.7
& 253.9
& 175.9 \\

& HIJP
& \best{13.5}
& \best{20.8}
& \best{36.2}
& \best{37.4}
& \best{27.0} \\

\cmidrule(l){2-7}
& \textbf{Speedup}
& $5.2\times$
& $6.0\times$
& $7.0\times$
& $6.8\times$
& $6.2\times$ \\

\midrule

% ======================== Reacher ========================

\multirow{3}{*}{Reacher}
& LeWM
& 153.6
& 296.0
& 566.6
& 581.9
& 399.5 \\

& HIJP
& \best{32.6}
& \best{45.4}
& \best{65.3}
& \best{69.0}
& \best{53.1} \\

\cmidrule(l){2-7}
& \textbf{Speedup}
& $4.7\times$
& $6.5\times$
& $8.7\times$
& $8.4\times$
& $7.1\times$ \\

\midrule

% ======================== PushT ========================

\multirow{3}{*}{PushT}
& LeWM
& 183.0
& 368.5
& 682.3
& 669.6
& 475.8 \\

& HIJP
& \best{14.9}
& \best{24.3}
& \best{41.4}
& \best{41.6}
& \best{30.6} \\

\cmidrule(l){2-7}
& \textbf{Speedup}
& $12.3\times$
& $15.2\times$
& $16.5\times$
& $16.1\times$
& $15.0\times$ \\

\midrule

% ======================== OGCube ========================

\multirow{3}{*}{OGCube}
& LeWM
& 110.5
& 180.9
& 298.2
& 297.7
& 221.8 \\

& HIJP
& \best{59.5}
& \best{70.5}
& \best{87.4}
& \best{90.8}
& \best{77.0} \\

\cmidrule(l){2-7}
& \textbf{Speedup}
& $1.9\times$
& $2.6\times$
& $3.4\times$
& $3.3\times$
& $2.8\times$ \\

\bottomrule
\end{tabular*}
\end{table}

The result shows that HIJP reduces planning time across all evaluated settings, achieving $1.9$--$16.5\times$ speedups over LeWM under matched replanning counts.
The average speedups are $6.2\times$, $7.1\times$, $15.0\times$, and $2.8\times$ on TwoRoom, Reacher, PushT, and OGCube, respectively.
Together with the results in Table~\ref{tab:main}, these findings show that HIJP achieves substantial planning speedups over LeWM while attaining higher average success rates on all four tasks.

\subsection{Ablations}
\label{sec:exp-ablation}

To systematically evaluate the contribution of each component, we conduct ablation studies on the training objectives, gradient propagation, hindsight goal relabeling, and the test-time planner.
For the training objectives, we remove one term at a time from Eq.~\ref{eq:total}: the inverse loss $\mathcal{L}_{\mathrm{inv}}$, the transition-consistency loss $\mathcal{L}_{\mathrm{cyc}}$, and the SIGReg regularizer $\mathcal{L}_{\mathrm{sigreg}}$.

We further evaluate two gradient-related variants. In $\operatorname{sg}[z_g]$, we stop gradients through the goal representation $z_g$. 
In the frozen-$g_\phi$ variant, we retain the latent predictor $g_\phi$ in the forward computation but block parameter updates from $\mathcal{L}{\mathrm{cyc}}$ while retaining those from $\mathcal{L}{\mathrm{pred}}$.
To evaluate the contribution of hindsight goal relabeling, we introduce a fixed-goal variant that always uses the latent representation of the final observation in each training window as the goal. Specifically, we sample an anchor step $t$ uniformly from $\{0,\ldots,T-2\}$, set $z_g=z_{T-1}$, and predict the one-step action $a_t$ from $(z_t,z_g)$. 

For the planner ablation, we keep the fully trained representation fixed and replace our amortized rollout with CEM search, thereby
isolating the effect of the planning mechanism. 
All other settings follow Section~\ref{sec:exp-setup}. 
We report results on Reacher, PushT, and OGCube in Table~\ref{tab:ablation}.
We exclude TwoRoom because all variants achieve near-100\% success rates on this task.
Furthermore, we examine sensitivity to the inverse and transition-consistency loss weights in Appendix~\ref{app:loss_weight_ablation}.

\begin{table*}[ht]
\centering
\caption{Success rates for HIJP and its ablation variants. Avg. reports the arithmetic mean across the four difficulty tiers for each task, and Overall Avg. averages all 12 task--tier combinations. The best and second-best results in each row are shown in \textbf{bold} and \underline{underlined text}.
}

\label{tab:ablation}

\footnotesize
\normalfont
\fontsize{7.75pt}{8.225pt}\selectfont
\setlength{\tabcolsep}{2pt}
\renewcommand{\arraystretch}{1.05} % 只影响行高
\setlength{\tabcolsep}{1.15pt}       % 只影响列间距

\begin{tabular*}{0.95\textwidth}{
    @{\extracolsep{\fill}}
    ll
    *{8}{c}
    @{}
}
\toprule
& & Full model
& \multicolumn{2}{c}{Training objectives}
& \multicolumn{2}{c}{Goal supervision}
& \multicolumn{2}{c}{Action Consistency}
& \multicolumn{1}{c}{Planner} \\
\cmidrule(lr){3-3}
\cmidrule(lr){4-5}
\cmidrule(lr){6-7}
\cmidrule(lr){8-9}
\cmidrule(lr){10-10}

Task
& Tier
& \textbf{HIJP}
& w/o $\mathcal{L}_{\mathrm{inv}}$
& w/o $\mathcal{L}_{\mathrm{sigreg}}$
& Endpoint goal
& $\operatorname{sg}[z_g]$
& w/o $\mathcal{L}_{\mathrm{cyc}}$
& Frozen $g_\phi$
& CEM \\
\midrule

\multirow{5}{*}{Reacher}
& Easy
& \best{48\%}
& 36\%
& 2\%
& 36\%
& \snd{40\%}
& 34\%
& \snd{40\%}
& \snd{40\%} \\

& Default
& \best{92\%}
& 72\%
& 0\%
& 66\%
& \snd{88\%}
& 78\%
& 84\%
& 70\% \\

& Hard
& \snd{92\%}
& 90\%
& 0\%
& 72\%
& 88\%
& 84\%
& 80\%
& \best{98\%} \\

& Extreme
& \best{94\%}
& 86\%
& 0\%
& 82\%
& \snd{92\%}
& \best{94\%}
& 84\%
& \best{94\%} \\

\cmidrule(l){2-10}
& \textbf{Avg.}
& \best{81.5\%}
& 71.0\%
& 0.5\%
& 64.0\%
& \snd{77.0\%}
& 72.5\%
& 72.0\%
& 75.5\% \\
\midrule

\multirow{5}{*}{PushT}
& Easy
& \best{90\%}
& 36\%
& 74\%
& 76\%
& \best{90\%}
& 84\%
& \snd{86\%}
& 78\% \\

& Default
& \best{80\%}
& 6\%
& 58\%
& 74\%
& \best{80\%}
& \snd{78\%}
& 72\%
& 74\% \\

& Hard
& \best{66\%}
& 0\%
& 34\%
& 50\%
& 48\%
& 54\%
& \snd{64\%}
& 44\% \\

& Extreme
& 42\%
& 0\%
& 20\%
& \snd{44\%}
& 38\%
& 22\%
& \best{46\%}
& 24\% \\

\cmidrule(l){2-10}
& \textbf{Avg.}
& \best{69.5\%}
& 10.5\%
& 46.5\%
& 61.0\%
& 64.0\%
& 59.5\%
& \snd{67.0\%}
& 55.0\% \\
\midrule

\multirow{5}{*}{OGCube}
& Easy
& \best{98\%}
& 80\%
& \best{98\%}
& 92\%
& 94\%
& \best{98\%}
& \best{98\%}
& \snd{96\%} \\

& Default
& \best{96\%}
& 56\%
& \best{96\%}
& \snd{88\%}
& \best{96\%}
& \best{96\%}
& \best{96\%}
& 80\% \\

& Hard
& \best{96\%}
& 46\%
& 90\%
& 88\%
& \snd{94\%}
& \snd{94\%}
& \snd{94\%}
& 58\% \\

& Extreme
& \best{96\%}
& 40\%
& 92\%
& 92\%
& \snd{94\%}
& \best{96\%}
& \best{96\%}
& 76\% \\

\cmidrule(l){2-10}
& \textbf{Avg.}
& \best{96.5\%}
& 55.5\%
& 94.0\%
& 90.0\%
& 94.5\%
& \snd{96.0\%}
& \snd{96.0\%}
& 77.5\% \\

\specialrule{0.08em}{0.4em}{0.3em}
\rowcolor{overallgray}
\multicolumn{2}{c}{\textbf{Overall Avg.}}
& \best{82.5\%}
& 45.7\%
& 47.0\%
& 71.7\%
& \snd{78.5\%}
& 76.0\%
& 78.3\%
& 69.3\% \\
\bottomrule
\end{tabular*}
\end{table*}

\begin{figure}[h]
\centering
\includegraphics[width=\linewidth]{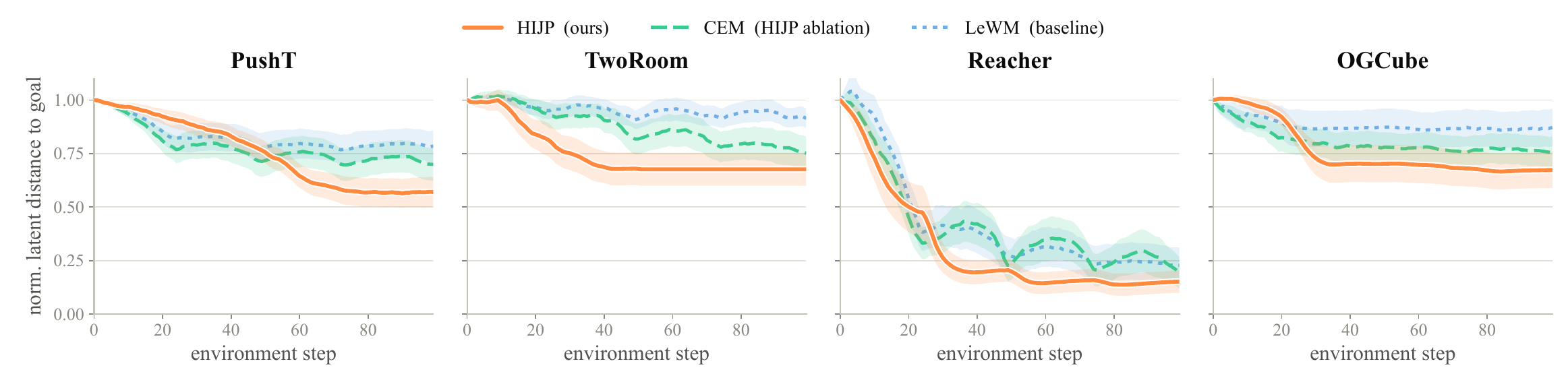}
\vspace{-1em}
\caption{ Normalized latent distance to the goal for HIJP, its CEM ablation, and LeWM across four tasks on the extreme tier.
Lines show means, and shaded regions indicate pointwise 95\% confidence intervals.
HIJP and its CEM ablation share the same latent space, while LeWM uses its own encoder. }
\label{fig:latent}
\vspace{-1.5em}
\end{figure}

%写作逻辑 ： 先说明消融要检验什么 → 报告总体下降 → 比较不同任务上的下降幅度 → 将差异联系到设计动机 → 最后讨论额外组件

The results show that removing $\mathcal{L}_{\mathrm{inv}}$ causes the largest overall decline, reducing average success from 82.5\% to 45.7\%.
Removing SIGReg also reduces average success substantially, to 47.0\%, with near-collapse on Reacher.
These results support the importance of action supervision and representation regularization for HIJP's control performance.
The two goal-supervision ablations examine different parts of our design choice.
Restricting goals to the final observation of each window reduces average success to 71.7\%, while stopping gradients through $z_g$ reduces it to 78.5\%.
This supports using multiple future observations as goals and allowing action-related gradients to shape goal representations.

The action consistency ablations examine the contribution of the transition-consistency objective.
Removing $\mathcal{L}_{\mathrm{cyc}}$ lowers average success to 76.0\%, with the clearest decreases on Reacher and PushT and little change on OGCube.
Blocking predictor updates from \(\mathcal{L}_{\mathrm{cyc}}\) lowers average success to 78.3\%, primarily affecting Reacher.
These results support the transition-consistency objective and its updates to the latent predictor, with gains varying across tasks.
Finally, replacing the inverse-guided rollout with CEM reduces average success to 69.3\% under the matched evaluation protocol, supporting the effectiveness of HIJP's learned planning mechanism.

%-------------------------
\subsection{Latent-Space Progress Analysis}
\label{sec:exp-qual}

HIJP shows its largest gains on long-horizon goals. We next examine how the latent distance to the goal evolves during execution.
We compare HIJP, its CEM ablation (Section~\ref{sec:exp-ablation}), and LeWM. 
The CEM ablation shares HIJP's learned representation and replaces HIJP's amortized inverse-guided rollout with CEM search, isolating the contribution of the planner.
On the extreme tier ($K{=}75$), we execute all methods in the original simulator on the same start--goal instances, following each method's replanning schedule, and track the normalized latent distance to the goal, $d_t=\lVert z_t-z_g\rVert/\lVert z_0-z_g\rVert$, where $z_t$ and $z_g$ are the encoded observation and goal.
After an episode succeeds, we hold its last recorded distance constant for the remaining steps. Figure~\ref{fig:latent} reports the mean and pointwise 95\% confidence intervals.

Figure~\ref{fig:latent} shows that HIJP reaches a lower final normalized latent distance than its CEM ablation on all four tasks.
On PushT, the methods exhibit similar initial progress, but HIJP continues to reduce the distance later in the episode as CEM largely plateaus.
On TwoRoom and OGCube, HIJP reaches a lower plateau, while on Reacher it exhibits fewer fluctuations after the initial decrease.
Compared with its CEM ablation, HIJP's executed trajectories end closer to the goal in the shared latent representation.

%-------------------------
\subsection{Smoothness of Executed Actions}
\label{sec:exp-smoothness}

\begin{wraptable}{r}{0.55\linewidth}
\vspace{-0.8\baselineskip}
\centering
\caption{ Action jerk on the extreme tier.  Ratio denotes each comparator's jerk divided by HIJP's. The lowest jerk in each environment is shown in \textbf{bold}.}
\label{tab:jerk-main}
\footnotesize
\setlength{\tabcolsep}{5pt}
\resizebox{\linewidth}{!}{%
\begin{tabular}{lccccc}
\toprule
& \textbf{HIJP}
& \multicolumn{2}{c}{CEM (ablation)}
& \multicolumn{2}{c}{LeWM} \\
\cmidrule(lr){2-2}
\cmidrule(lr){3-4}
\cmidrule(lr){5-6}
Environment
& Jerk $\downarrow$
& Jerk $\downarrow$
& Ratio $\uparrow$
& Jerk $\downarrow$
& Ratio $\uparrow$ \\
\midrule
PushT
& \textbf{0.048} & 0.697 & 14.7$\times$
& 0.724 & 15.2$\times$ \\
TwoRoom
& \textbf{0.146} & 2.503 & 17.1$\times$
& 2.558 & 17.5$\times$ \\
Reacher
& \textbf{0.046} & 1.222 & 26.4$\times$
& 1.134 & 24.5$\times$ \\
OGCube
& \textbf{0.099} & 2.065 & 20.9$\times$
& 2.304 & 23.3$\times$ \\
\bottomrule
\end{tabular}%
}
\vspace{-0.6\baselineskip}
\end{wraptable}

We additionally evaluate the temporal smoothness of executed actions.
We quantify this with action jerk, the mean $\ell_2$ norm of the second finite difference of the executed action sequence, $\tfrac{1}{T-2}\sum_t \lVert \mathbf{a}_t - 2\mathbf{a}_{t-1} + \mathbf{a}_{t-2}\rVert_2$ (lower is smoother), following the convention of \citet{nguyen2026latentgeometrysearchamortizing}. To attribute any difference to the planner rather than the representation, we compare HIJP against its CEM ablation, which shares the same learned representation and differs only in the planning rule, and include LeWM as a reference.
Table~\ref{tab:jerk-main} reports the extreme tier ($K{=}75$); the metric definition and full per-tier results are in Appendix~\ref{app:smoothness}.

The results show that HIJP produces smoother executed actions across all four tasks, with action jerk $14.7$--$26.4\times$ lower than its CEM ablation and $15.2$--$24.5\times$ lower than LeWM.
These results indicate that HIJP's amortized rollout improves the temporal smoothness of executed controls in addition to its goal-reaching performance and planning efficiency.

\section{Conclusion}
\label{section:conclusion}
We introduced Hindsight Inverse JEPA Planning (HIJP), a search-free framework for goal-conditioned visual control. HIJP samples future observations as hindsight goals to enrich inverse-dynamics supervision and allows training gradients to update goal representations. It further feeds predicted actions into a latent predictor model and constrains the resulting latent transitions through a consistency objective. At planning time, the inverse and latent predictor models alternate to construct multi-step action plans without online search. Across four visual control tasks and multiple evaluation configurations, HIJP achieves the highest aggregate goal-reaching performance among the evaluated methods while reducing planning time compared to a search-based world-model planner. Under the most challenging setting, its executed actions are also smoother than those produced by CEM using the same representations. 
% At planning time, HIJP alternates between inverse action prediction and latent prediction to construct multi-step action plans without online search.
Together, these results suggest that aligning goal-conditioned action learning with the latent transitions used during planning provides an effective route toward efficient, search-free visual control.
% Our evaluation is limited to offline-trained control in simulated environments with fixed planning horizons and replanning schedules; robustness in real-world settings remains untested. 
% Future work will explore adaptive horizons and replanning schedules, as well as real-world deployment.

% 我们提出HIJP，这是针对目标条件视觉控制的面搜索规划方法。 它利用hindsight重标记的方法随机把未来观测作为hindsight goal， 增强逆动力学的监督，并允许梯度更新目标； 同时，它还可以通过把预测的action 输入到前向模型中来预测动作引起的潜在状态转移。在测试阶段，逆向模型与前向模型交替运作，无需在线搜索即可构建多步动作规划。在4项视觉控制任务及多种评估配置下，我们的方法在所有基线中展现出了最优的综合目标达成性能，且相比于基于搜索的世界模型，规划器可以有更短的规划时间与更强的动作平滑性。消融实验验证了Hindsight sampling 以，目标表征的梯度传播和 动作预测转移一致性这几项机制的有效性。结果表明，将目标条件动作预测与前向潜在动力学相结合，是实现高效视觉控制的一种极具前景的方法。未来的研究将讨论自适应规划时域以及在现实世界中的控制应用。

\section{Limitations} 
Our evaluation focuses on tasks with a limited temporal span, using fixed planning horizons and replanning schedules.
Extending HIJP to substantially longer-horizon tasks remains an important direction.
Future work will explore hierarchical subgoal generation, adaptive planning horizons, and dynamic replanning schedules.
These extensions aim to decompose longer tasks into shorter goal-conditioned segments and adapt planning and execution intervals to the requirements of each segment.

\bibliography{iclr2027_conference}

@inproceedings{bo2026diagram,
  title={Diagram-MMU: A Multi-modal Benchmark for Scientific Diagrams},
  author={Bo, Weihao and Zhang, Shan and Sun, Yanpeng and Liu, Jie and Yao, Yongke and Du, Jinhao and He, Wei and Zou, Kai and Li, Zechao and Wang, Jingdong},
  booktitle={European Conference on Computer Vision},
  pages={524--543},
  year={2026},
  organization={Springer}
}

@article{tang2025artemis,
  title={Artemis: Structured Visual Reasoning for Perception Policy Learning},
  author={Tang, Wei and Sun, Yanpeng and Zhang, Shan and Bo, Weihao and Li, Xiaofan and Koniusz, Piotr and Li, Wei and Zhao, Na and Li, Zechao},
  journal={arXiv preprint arXiv:2512.01988},
  year={2025}
}

@misc{zeng2026dtagtdirectiontopologyawaregraph,
      title={DTA-GT: Direction- and Topology-Aware Graph Transformer for Neural Network Representation Learning}, 
      author={Yuxiang Zeng and Kun Xie and Yang Wang and Jigang Wen and Xiaocan Li and Guangxing Zhang and Gaogang Xie and Jiannong Cao},
      year={2026},
      eprint={2610.16822},
      archivePrefix={arXiv},
      primaryClass={cs.LG},
      url={https://arxiv.org/abs/2610.16822}, 
}

@inproceedings{dosovitskiy2021vit,
title={An Image is Worth 16x16 Words: Transformers for Image Recognition at Scale},
author={Alexey Dosovitskiy and Lucas Beyer and Alexander Kolesnikov and Dirk Weissenborn and Xiaohua Zhai and Thomas Unterthiner and Mostafa Dehghani and Matthias Minderer and Georg Heigold and Sylvain Gelly and Jakob Uszkoreit and Neil Houlsby},
booktitle={International Conference on Learning Representations},
year={2021},
url={https://openreview.net/forum?id=YicbFdNTTy}
}

@InProceedings{Peebles2023dit,
    author    = {Peebles, William and Xie, Saining},
    title     = {Scalable Diffusion Models with Transformers},
    booktitle = {Proceedings of the IEEE/CVF International Conference on Computer Vision (ICCV)},
    month     = {October},
    year      = {2023},
    pages     = {4195-4205}
}

@article{maes2026leworldmodel,
  title={Leworldmodel: Stable end-to-end joint-embedding predictive architecture from pixels},
  author={Maes, Lucas and Lidec, Quentin Le and Scieur, Damien and LeCun, Yann and Balestriero, Randall},
  journal={arXiv preprint arXiv:2603.19312},
  year={2026}
}

@inproceedings{
zhou2025dinowm,
title={{DINO}-{WM}: World Models on Pre-trained Visual Features enable Zero-shot Planning},
author={Gaoyue Zhou and Hengkai Pan and Yann LeCun and Lerrel Pinto},
booktitle={Forty-second International Conference on Machine Learning},
year={2025},
url={https://openreview.net/forum?id=D5RNACOZEI}
}

@article{balestriero2025lejepa,
  title={Lejepa: Provable and scalable self-supervised learning without the heuristics},
  author={Balestriero, Randall and LeCun, Yann},
  journal={arXiv preprint arXiv:2511.08544},
  year={2025}
}

@inproceedings{
ghosh2021learning,
title={Learning to Reach Goals via Iterated Supervised Learning},
author={Dibya Ghosh and Abhishek Gupta and Ashwin Reddy and Justin Fu and Coline Manon Devin and Benjamin Eysenbach and Sergey Levine},
booktitle={International Conference on Learning Representations},
year={2021},
url={https://openreview.net/forum?id=rALA0Xo6yNJ}
}

@inproceedings{
sobal2025stresstesting,
title={Stress-Testing Offline Reward-Free Reinforcement Learning: A Case for Planning with Latent Dynamics Models},
author={Vlad Sobal and Wancong Zhang and Kyunghyun Cho and Randall Balestriero and Tim G. J. Rudner and Yann LeCun},
booktitle={7th Robot Learning Workshop: Towards Robots with Human-Level Abilities},
year={2025},
url={https://openreview.net/forum?id=jON7H6A9UU}
}

@misc{ivashkov2026sensorimotorworldmodelsperception,
      title={Sensorimotor World Models: Perception for Action via Inverse Dynamics}, 
      author={Petr Ivashkov and Randall Balestriero and Bernhard Schölkopf},
      year={2026},
      eprint={2606.20104},
      archivePrefix={arXiv},
      primaryClass={cs.LG},
      url={https://arxiv.org/abs/2606.20104}, 
}

@misc{zhang2026deltajepalearningactionsensitiveworld,
      title={Delta-JEPA: Learning Action-Sensitive World Models via Latent Difference Decoding}, 
      author={Zhenghao Zhang and Yuanxiang Wang and Zhenyu Guan and Yujia Yang and Bingkang Shi and Tianyu Zong and Hongzhu Yi and Guoqing Chao and Xingchen Chen and Tiankun Yang and Chenxi Bao and Tao Yu and Jingjing Zhou and Jungang Xu},
      year={2026},
      eprint={2606.31232},
      archivePrefix={arXiv},
      primaryClass={cs.AI},
      url={https://arxiv.org/abs/2606.31232}, 
}

@misc{vjepa2Mido2025,
      title={V-JEPA 2: Self-Supervised Video Models Enable Understanding, Prediction and Planning}, 
      author={Mido Assran and Adrien Bardes and David Fan and Quentin Garrido and Russell Howes and Mojtaba and Komeili and Matthew Muckley and Ammar Rizvi and Claire Roberts and Koustuv Sinha and Artem Zholus and Sergio Arnaud and Abha Gejji and Ada Martin and Francois Robert Hogan and Daniel Dugas and Piotr Bojanowski and Vasil Khalidov and Patrick Labatut and Francisco Massa and Marc Szafraniec and Kapil Krishnakumar and Yong Li and Xiaodong Ma and Sarath Chandar and Franziska Meier and Yann LeCun and Michael Rabbat and Nicolas Ballas},
      year={2025},
      eprint={2506.09985},
      archivePrefix={arXiv},
      primaryClass={cs.AI},
      url={https://arxiv.org/abs/2506.09985}, 
}

@article{sun2019policy,
  title={Policy continuation with hindsight inverse dynamics},
  author={Sun, Hao and Li, Zhizhong and Liu, Xiaotong and Zhou, Bolei and Lin, Dahua},
  journal={Advances in Neural Information Processing Systems},
  volume={32},
  year={2019}
}

@misc{nguyen2026latentgeometrysearchamortizing,
      title={Latent Geometry Beyond Search: Amortizing Planning in World Models}, 
      author={Hoang Nguyen and Xiaohao Xu and Xiaonan Huang},
      year={2026},
      eprint={2605.08732},
      archivePrefix={arXiv},
      primaryClass={cs.RO},
      url={https://arxiv.org/abs/2605.08732}, 
}

@misc{sun2026intactisomorphicintenttoactionlearning,
      title={INTACT: Isomorphic Intent-to-Action Learning for Search-Free World Models}, 
      author={Junhan Sun and Hao Zhao and Guofeng Zhang},
      year={2026},
      eprint={2607.26056},
      archivePrefix={arXiv},
      primaryClass={cs.RO},
      url={https://arxiv.org/abs/2607.26056}, 
}

@misc{seo2026acidactionconsistencyinverse,
      title={ACID: Action Consistency via Inverse Dynamics for Planning with World Models}, 
      author={Gawon Seo and Dongwon Kim and Suha Kwak},
      year={2026},
      eprint={2607.02403},
      archivePrefix={arXiv},
      primaryClass={cs.RO},
      url={https://arxiv.org/abs/2607.02403}, 
}

@article{terver2025drives,
  title={What drives success in physical planning with joint-embedding predictive world models?},
  author={Terver, Basile and Yang, Tsung-Yen and Ponce, Jean and Bardes, Adrien and LeCun, Yann},
  journal={arXiv preprint arXiv:2512.24497},
  year={2025}
}

@article{gao2026fast,
  title={Fast LeWorldModel},
  author={Gao, Yuntian and Xu, Xiangyu},
  journal={arXiv preprint arXiv:2606.26217},
  year={2026}
}

@misc{liu2026physicallygroundedjepaworld,
      title={Toward Physically Grounded JEPA World Models for Goal-Conditioned Robotic Planning}, 
      author={Muyuan Liu and Yue Huang and Zheng Liang and Xiang Gao},
      year={2026},
      eprint={2609.03565},
      archivePrefix={arXiv},
      primaryClass={cs.RO},
      url={https://arxiv.org/abs/2609.03565}, 
}

@article{wang2026decision,
  title={Decision-Metric Alignment in Latent World Models: Diagnostics and Action-Conditioned Objectives for MPC Planning},
  author={Wang, Jiawei and Rui, Ke and Zuo, Yushen and Feng, Yichun and Li, Minglei},
  journal={arXiv preprint arXiv:2608.18746},
  year={2026}
}

@misc{andrychowicz2017hindsight,
      title={Hindsight Experience Replay}, 
      author={Marcin Andrychowicz and Filip Wolski and Alex Ray and Jonas Schneider and Rachel Fong and Peter Welinder and Bob McGrew and Josh Tobin and Pieter Abbeel and Wojciech Zaremba},
      year={2018},
      eprint={1707.01495},
      archivePrefix={arXiv},
      primaryClass={cs.LG},
      url={https://arxiv.org/abs/1707.01495}, 
}

@article{ghosh2019learning,
  title={Learning to reach goals via iterated supervised learning},
  author={Ghosh, Dibya and Gupta, Abhishek and Reddy, Ashwin and Fu, Justin and Devin, Coline and Eysenbach, Benjamin and Levine, Sergey},
  journal={arXiv preprint arXiv:1912.06088},
  year={2019}
}

@article{paster2020planning,
  title={Planning from pixels using inverse dynamics models},
  author={Paster, Keiran and McIlraith, Sheila A and Ba, Jimmy},
  journal={arXiv preprint arXiv:2012.02419},
  year={2020}
}

@article{schafer2026does,
  title={When does predictive inverse dynamics outperform behavior cloning?},
  author={Sch{\"a}fer, Lukas and Choudhury, Pallavi and Lemkhenter, Abdelhak and Lovett, Chris and Nath, Somjit and Fran{\c{c}}a, Luis and de Mendon{\c{c}}a, Matheus Ribeiro Furtado and Lamb, Alex and Islam, Riashat and Sen, Siddhartha and others},
  journal={arXiv preprint arXiv:2601.21718},
  year={2026}
}

@article{morin2026sample,
  title={On the Sample Efficiency of Inverse Dynamics Models for Semi-Supervised Imitation Learning},
  author={Morin, Sacha and Byeon, Moonsub and Jolicoeur-Martineau, Alexia and Lachapelle, S{\'e}bastien},
  journal={arXiv preprint arXiv:2602.02762},
  year={2026}
}

@article{wang2026prism,
  title={PRISM: PRior-guided Imagination Sampling in world Models},
  author={Wang, Yuhai and Xia, Jiawei and Zhou, Rongxuan and Hu, Xiao and Shi, Yongliang and Du, Jing and Ye, Yang},
  journal={arXiv preprint arXiv:2606.07974},
  year={2026}
}

@article{cheng2026sage,
  title={SAGE: Subgoal-Conditioned Action Generation for Latent World Model Planning},
  author={Cheng, Letian and Zhang, Qi and Wang, Yisen},
  journal={arXiv preprint arXiv:2607.17973},
  year={2026}
}

@article{masip2026ff,
  title={FF-JEPA: Long-Horizon Planning in World Models with Latent Planners},
  author={Masip, Sergi and Swinnen, Jonathan and Hu, Yutong and Detry, Renaud and Tuytelaars, Tinne},
  journal={arXiv preprint arXiv:2606.09311},
  year={2026}
}

@article{sommer2026reinforced,
  title={Reinforced Planning with Latent World Models},
  author={Sommer, Armin and Schilling, Jannik},
  journal={arXiv preprint arXiv:2608.18669},
  year={2026}
}

@inproceedings{tian2025predictive,
  title={Predictive inverse dynamics models are scalable learners for robotic manipulation},
  author={Tian, Yang and Yang, Sizhe and Zeng, Jia and Wang, Ping and Lin, Dahua and Dong, Hao and Pang, Jiangmiao},
  booktitle={International Conference on Learning Representations},
  volume={2025},
  pages={92033--92052},
  year={2025}
}

@inproceedings{
xie2025latent,
title={Latent Diffusion Planning for Imitation Learning},
author={Amber Xie and Oleh Rybkin and Dorsa Sadigh and Chelsea Finn},
booktitle={Forty-second International Conference on Machine Learning},
year={2025},
url={https://openreview.net/forum?id=vhACnRfuYh}
}

@article{ajay2022conditional,
  title={Is conditional generative modeling all you need for decision-making?},
  author={Ajay, Anurag and Du, Yilun and Gupta, Abhi and Tenenbaum, Joshua and Jaakkola, Tommi and Agrawal, Pulkit},
  journal={arXiv preprint arXiv:2211.15657},
  year={2022}
}

@article{huang2026leflow,
  title={LeFlow: Generative Latent Flow Planning for World Models},
  author={Huang, Hsiang-Wei and Shangguan, Jianxu and Lu, Junbin and Hwang, Jenq-Neng},
  journal={arXiv preprint arXiv:2608.24855},
  year={2026}
}

@article{rakhimov2026qantara,
  title={Qantara: Bridge-Flow Training for Multi-Paradigm JEPA Control},
  author={Rakhimov, Ruslan and Bredis, George and Maksyuta, Yuriy and Gavrilov, Daniil},
  journal={arXiv preprint arXiv:2607.04978},
  year={2026}
}

@article{bai2026temporal,
  title={Temporal-Distance JEPA: Plan-Aware Representation Learning for Latent World Model Predictive Control},
  author={Bai, Jiaxin and Xiong, Jiaxuan},
  journal={arXiv preprint arXiv:2607.25337},
  year={2026}
}

@misc{zhang2025efficientimagegoalnavigationrepresentative,
      title={Efficient Image-Goal Navigation with Representative Latent World Model}, 
      author={Zhiwei Zhang and Hui Zhang and Kaihong Huang and Chenghao Shi and Huimin Lu},
      year={2025},
      eprint={2511.11011},
      archivePrefix={arXiv},
      primaryClass={cs.RO},
      url={https://arxiv.org/abs/2511.11011}, 
}

@article{rubinstein1999cross,
  title={The cross-entropy method for combinatorial and continuous optimization},
  author={Rubinstein, Reuven},
  journal={Methodology and computing in applied probability},
  volume={1},
  number={2},
  pages={127--190},
  year={1999},
  publisher={Springer}
}

@inproceedings{williams2017information,
  title={Information theoretic MPC for model-based reinforcement learning},
  author={Williams, Grady and Wagener, Nolan and Goldfain, Brian and Drews, Paul and Rehg, James M and Boots, Byron and Theodorou, Evangelos A},
  booktitle={2017 IEEE international conference on robotics and automation (ICRA)},
  pages={1714--1721},
  year={2017},
  organization={IEEE}
}

@inproceedings{pinneri2021sample,
  title={Sample-efficient cross-entropy method for real-time planning},
  author={Pinneri, Cristina and Sawant, Shambhuraj and Blaes, Sebastian and Achterhold, Jan and Stueckler, Joerg and Rolinek, Michal and Martius, Georg},
  booktitle={Conference on Robot Learning},
  pages={1049--1065},
  year={2021},
  organization={PMLR}
}
\bibliographystyle{iclr2027_conference}

\clearpage
\appendix

\section*{Appendix}

\section{Dataset Details and Protocol}
\label{app:datasets}

\begin{figure}[h]
\centering
\includegraphics[width=\linewidth]{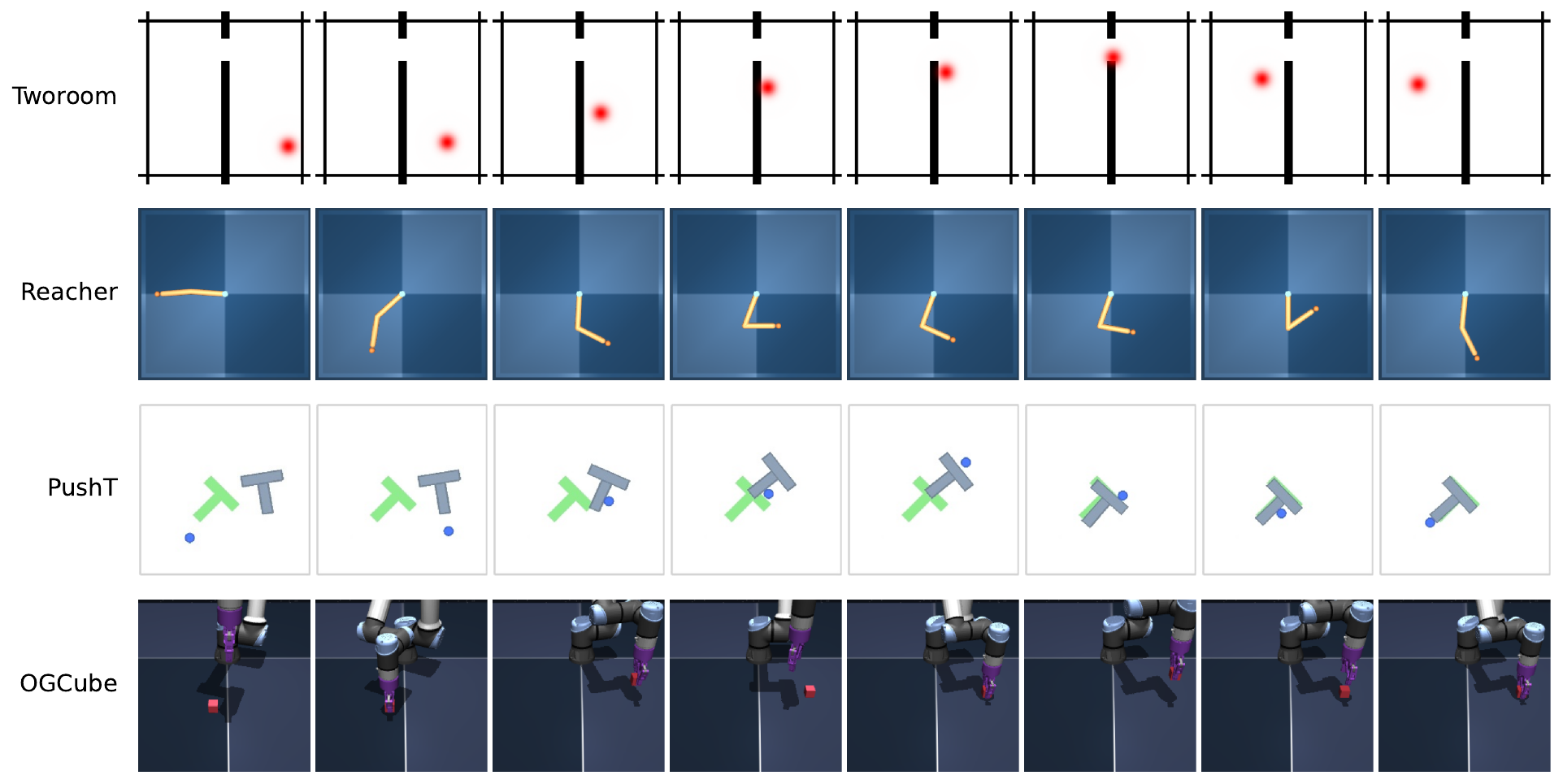}
\caption{Task overview. Each row shows one representative trajectory per benchmark, sampled uniformly in time from its start (leftmost) to its end (rightmost).}
\label{fig:tasks}
\end{figure}

We evaluate on four continuous-control benchmarks: \textbf{Tworoom}, \textbf{Reacher}, \textbf{PushT}, and \textbf{OGBench-Cube} (hereafter OGCube), which together span $2$D navigation, low-dimensional continuous control, planar object pushing, and robotic cube manipulation; Figure~\ref{fig:tasks} shows a representative trajectory from each. We do not collect any trajectories ourselves: we reuse the full-scale offline datasets released by LeWM~\citep{maes2026leworldmodel} without modification.

\subsection{Datasets}

\begin{figure}[t]
\centering
\includegraphics[width=\linewidth]{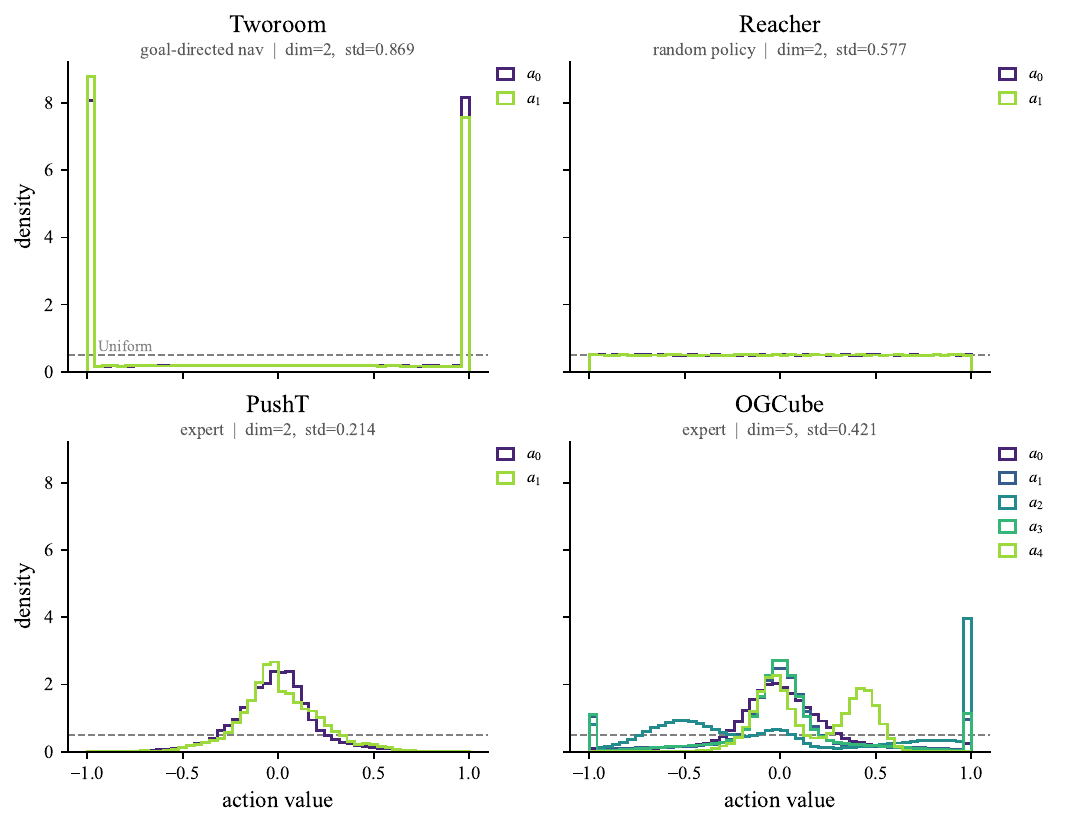}
\caption{Per-dimension marginal action distributions for each dataset. The dashed line marks the density of a uniform distribution over $[-1, 1]$.} 
% The behavior policy (subtitle) and per-dataset action standard deviation are shown above each panel; all actions are z-score standardized before training.}

\label{fig:action_dist}
\end{figure}

The four datasets are gathered under different behavior policies: a goal-directed navigation policy for Tworoom, a near-uniform random policy for Reacher, and expert policies for PushT and OGCube. This heterogeneity is directly visible in their action distributions (Figure~\ref{fig:action_dist}): Reacher actions are essentially uniform over $[-1, 1]$, Tworoom actions saturate at the $\pm 1$ boundaries, whereas PushT and OGCube produce concentrated, structured actions.

Each dataset is stored as a single HDF5 file under a shared schema: RGB observations \texttt{pixels} of shape $(N, 224, 224, 3)$ in uint8, a continuous \texttt{action} array of shape $(N, A)$, and per-episode boundaries encoded by \texttt{ep\_offset} and \texttt{ep\_len}, so that a sampled window never straddles two episodes. 
Table~\ref{tab:datasets} summarizes the resulting statistics.

\begin{table}[h]
\centering
\caption{Statistics of the four datasets (full versions, as used in all experiments). Frames denotes the total number of transitions; Traj.\ length reports the minimum, mean, and maximum episode lengths.}
\label{tab:datasets}
\setlength{\tabcolsep}{5pt}
\resizebox{\linewidth}{!}{%
\begin{tabular}{llccccc}
\toprule
Dataset & Simulator & Action dim $A$ & \#Episodes & Traj.\ length (min/mean/max) & \#Frames \\
\midrule
Tworoom & $2$D navigation   & $2$ & $10{,}000$ & $31\,/\,92\,/\,101$    & $920{,}809$   \\
Reacher & DMControl Reacher & $2$ & $10{,}000$ & $201\,/\,201\,/\,201$  & $2{,}010{,}000$ \\
PushT   & PushT             & $2$ & $18{,}685$ & $49\,/\,125\,/\,246$   & $2{,}336{,}736$ \\
OGCube  & OGBench Cube      & $5$ & $10{,}000$ & $201\,/\,201\,/\,201$  & $2{,}010{,}000$ \\
\bottomrule
\end{tabular}%
}
\end{table}

\paragraph{Observations and actions.} All observations are $224\times224$ RGB frames, which we feed to the encoder at full resolution. The action space is a $2$D command for Tworoom, Reacher, and PushT, and a $5$D vector (end-effector pose and gripper) for OGCube. Reacher and OGCube episodes have a fixed length of $201$ steps, whereas Tworoom and PushT episodes vary in length because they terminate upon task completion.

\paragraph{Action normalization.} Before training, every non-pixel input---the actions, and (where used) proprioception and state---is standardized per dimension by its dataset statistics, i.e., a z-score transform to zero mean and unit variance, applied identically to all four tasks. Figure~\ref{fig:action_dist} plots the corresponding \emph{raw} (pre-normalization) action distributions, which differ markedly across tasks because of their different behavior policies.

\paragraph{Training samples and monitoring split.}
Following the data-loading and window-level splitting procedure
of the LeWM implementation~\citep{maes2026leworldmodel}, we enumerate all valid windows within each trajectory and randomly assign 90\% to training and 10\% to validation.
Each window contains $W$ observations sampled at intervals of
$k$ raw environment steps, together with aligned action chunks.
The loader requires $Wk$ consecutive raw time-step records, while
the first and last sampled observations are separated by $(W-1)k$
environment steps.
We replace NaN action entries with zeros during training rather
than dropping the window.

\subsection{Evaluation Protocol}
\label{app:eval-protocol}

We follow the dataset-based evaluation procedure of the LeWM
implementation \citep{maes2026leworldmodel}. For each evaluation episode, we sample a start state from an offline trajectory and define the goal as the environment state $K$ steps ahead (at index $\texttt{start}+K$). The agent acts for at most $B$ steps. We count an episode as successful only if the simulator reports task completion within this budget Table~\ref{tab:success-criteria}; latent-space distance is not used to determine success.

\begin{table}[t]
\centering
\small
\caption{Environment-defined success criteria.}
\label{tab:success-criteria}
\begin{tabular}{@{}lp{0.72\linewidth}@{}}
\toprule
Task & Success condition \\
\midrule
Tworoom & Agent-to-goal position distance $<16$ in environment coordinates. \\
Reacher & Absolute error of every joint angle $<0.05$ radians relative to the target configuration. \\
PushT & Euclidean error over agent and block positions $<20$, and wrapped block orientation error $<\pi/9$. \\
OGCube & Target cube-to-goal position distance $\leq 0.04\,\mathrm{m}$. \\
\bottomrule
\end{tabular}
\end{table}

\begin{figure}[t]
\centering
\includegraphics[width=\linewidth]{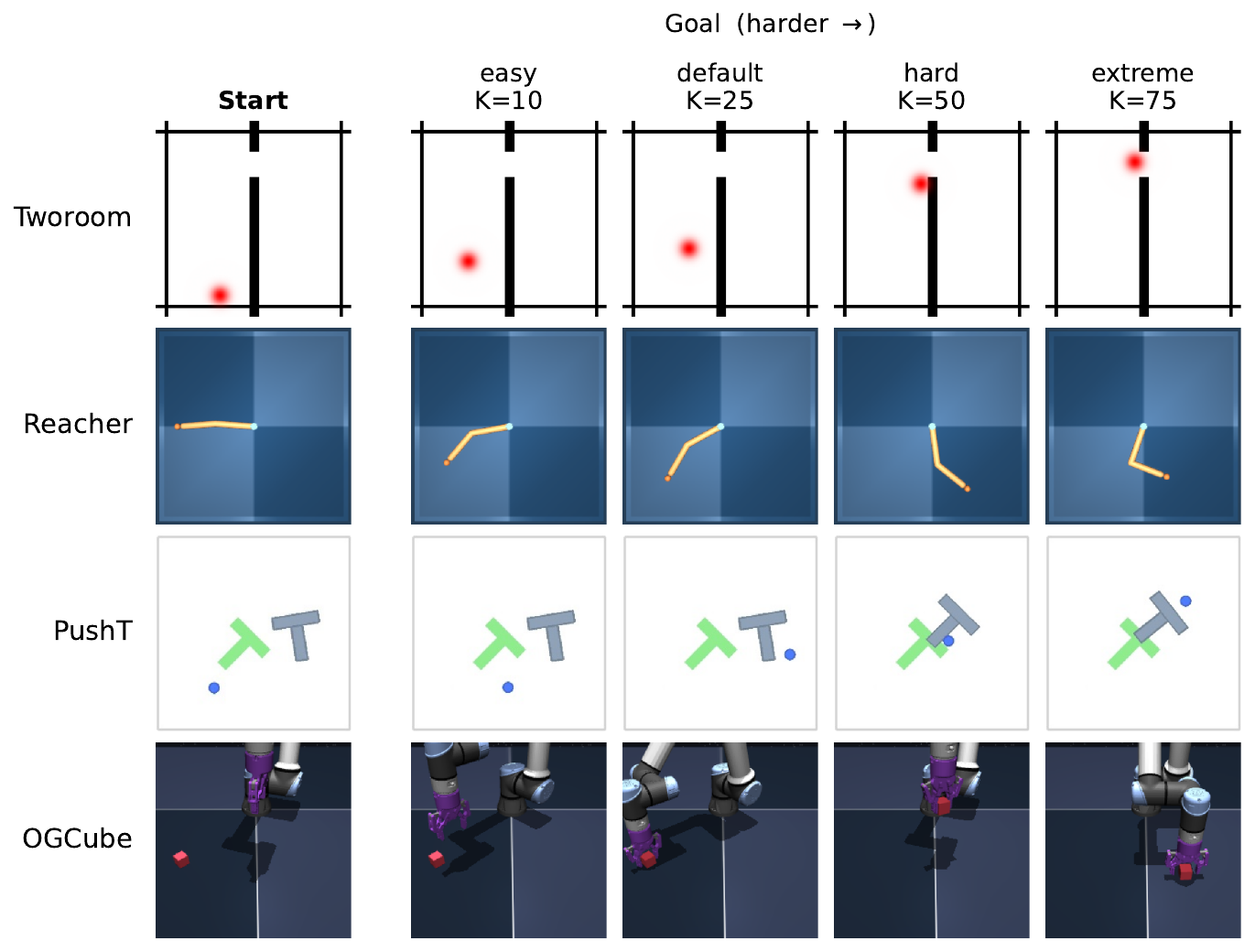}
\caption{Visualization of the four difficulty tiers. For each task (row), we fix a
start observation (leftmost column) and show the goal observation placed $K$ steps
ahead of it, one per tier. A larger offset $K$ moves the goal farther along the
trajectory and demands a longer planning horizon.}
\label{fig:difficulty}
\end{figure}

By varying the goal offset $K$ and the step budget $B$, we obtain four difficulty tiers (Table~\ref{tab:difficulty}; Figure~\ref{fig:difficulty}), ranging from \emph{easy} ($K{=}10$) to \emph{extreme} ($K{=}75$): a larger $K$ yields a longer-horizon goal. The extreme tier is the most challenging setting and is where our method attains its largest margin over the baselines.

\begin{table}[t]
\centering
\caption{Difficulty tiers used in evaluation. Each tier is fully specified by the
goal offset $K$ (how many steps ahead of the start the goal is placed) and the
step budget $B$ (the maximum number of environment steps allowed).}
\label{tab:difficulty}
\begin{tabular}{lcc}
\toprule
Tier & Goal offset $K$ & Step budget $B$ \\
\midrule
easy    & $10$ & $25$  \\
default & $25$ & $50$  \\
hard    & $50$ & $100$ \\
extreme & $75$ & $100$ \\
\bottomrule
\end{tabular}
\end{table}

\subsection{CEM Planning Protocol}
\label{app:cem-protocol}

CEM plays several roles in our experiments. It is used as the
test-time optimizer for the CEM-based world-model baselines, as the computational reference in the planning-efficiency experiment, and as a controlled planner ablation in which HIJP's amortized action
generation is replaced by search over the same learned representation and forward dynamics. We therefore use one shared CEM configuration throughout all experiments.

At every replanning point, CEM maintains a factorized Gaussian distribution over action sequences of length $H=5$ model steps.
Each model-level action contains a chunk of $k=5$ environment actions, so every candidate sequence covers $Hk=25$ environment steps. The sampling distribution is initialized with zero mean and sampling scale $1.0$.

At each optimization iteration, CEM samples $N=300$ candidate action sequences and rolls each candidate forward through the corresponding latent dynamics model. 
Candidates are ranked by the squared distance between their terminal predicted latent and the encoded goal,

\begin{equation}
    J_{\mathrm{goal}}(\mathbf{a}_{0:H-1})
    =
    \left\|
        \hat{z}_{H}(\mathbf{a}_{0:H-1})-z_g
    \right\|_2^2 .
\end{equation}

The best $30$ candidates are retained as elites and are used to refit the sampling mean and scale. This procedure is repeated for $I=30$ iterations.
Consequently, every CEM replanning call evaluates
\begin{equation}
    NI = 300 \times 30 = 9{,}000
\end{equation}
candidate action sequences. Since every sequence contains $H=5$ latent transitions, this corresponds to approximately
\begin{equation}
    NIH = 300 \times 30 \times 5 = 45{,}000
\end{equation}
latent transition predictions per environment and replanning call.
These operations are evaluated in parallel on the GPU; therefore, the sequence count should not be interpreted as an equivalent wall-clock slowdown.

All CEM comparisons use the same planning horizon, action chunking, interaction budget, and replanning cadence as HIJP. Both planners generate $H=5$ model-level actions and execute $R=5$ model steps, equivalent to $Rk=25$ environment steps, before observing the environment and replanning. Under the interaction budgets $B\in\{25,50,100,100\}$, the maximum numbers of replanning calls are therefore $1$, $2$, $4$, and $4$ for the easy, default, hard, and extreme tiers, respectively.

For the main LeWM baseline, CEM searches using LeWM's own encoder and forward model. In the controlled CEM ablation, it instead searches using HIJP's encoder and forward model while completely disabling the inverse-dynamics action generator. This controlled comparison changes only the test-time planning mechanism and therefore isolates the effect of amortized inverse-guided planning.

\begin{table}[t]
\centering
\small
\caption{
CEM configuration shared across all CEM-based comparisons.}
\label{tab:cem-config}
\begin{tabular}{lc}
\toprule
Hyperparameter & Value \\
\midrule
Planning horizon $H$              & $5$ model steps \\
Action-chunk length $k$           & $5$ environment steps \\
Execution stride $R$              & $5$ model steps \\
Environment steps per replan $Rk$ & $25$ \\
Candidate sequences $N$           & $300$ \\
Optimization iterations $I$       & $30$ \\
Elite sequences                   & $30$ \\
Initial sampling scale            & $1.0$ \\
Sequence evaluations per replan   & $9{,}000$ \\
Latent transitions per replan     & $45{,}000$ \\
\bottomrule
\end{tabular}
\end{table}

% \section{Baselines Details}
% \label{app:baselines}
% \input{text/app_baselines}

\section{Implementation Details}
\label{app:impl}
%% 这个是impliment的正文
This appendix provides a concrete instantiation of the method.
We first specify the temporal discretization and the architecture of each component, then describe the training procedure and test-time planning protocol.
Unless stated otherwise, the architecture and temporal settings are shared across all four tasks.

\subsection{Temporal Discretization and Data Flow}
\label{app:temporal-discretization}

The environments expose atomic actions $u_s\in\mathbb{R}^{d_a}$ at raw environment step $s$. The world model operates at a coarser temporal resolution by grouping $k$ consecutive atomic actions into one model-level action chunk,
$\mathbf a_t=\left(u_{tk},u_{tk+1},\ldots,u_{tk+k-1}\right) \in\mathbb{R}^{k d_a}.$

We use $k=5$ for every task.
Consequently, one transition in the learned latent dynamics corresponds to five raw environment steps. The raw action dimension is $d_a=2$ for Tworoom, Reacher, and PushT, and $d_a=5$ for OGCube. 
The effective model-level action dimension is therefore $d_{\mathbf a}=k d_a\in\{10,25\}$.

A training sample contains $W$ temporally subsampled observations, together with their aligned model-level action chunks. $ k=5$ raw environment steps separate consecutive observations. Thus, the first and final observations in a window are separated by $(W-1)k$ raw environment steps. 

Before being passed to the model, images are converted to floating-point, normalized using ImageNet statistics, and resized to $224\times224$. 
Following the LeWM preprocessing procedure, continuous actions are standardized per dimension using statistics computed from the training dataset. 
At execution time, predicted action chunks are mapped back to
the environment action space using the inverse transformation.

\subsection{Architecture}
\label{app:architecture}

The implementation contains a visual encoder $f_\theta$, an action encoder $\psi_\omega$, a latent predictor $g_\phi$, and a hindsight-conditioned inverse-dynamics head $h_\xi$. 
The action encoder is absorbed into the notation of $g_\phi$ in the main text and is made explicit here.

\paragraph{Visual encoder $f_\theta$.} 
We instantiate $f_\theta$ as a ViT-Tiny\citep{dosovitskiy2021vit} trained from scratch, without a pretrained backbone or an external visual representation.
It contains $12$ Transformer layers, $3$ attention heads, and a hidden width of $192$.
Each $224\times224$ observation is divided into patches of size $14\times14$.
We take the final\texttt{[CLS]} token and pass it through a projection MLP $192\!\rightarrow\!2048\!\rightarrow\!192$.
Batch normalization is applied at the hidden layer of the projection head.
The resulting vector is the latent representation $z_t=f_\theta(o_t)\in\mathbb{R}^{192}$.

\paragraph{Action encoder $\psi_\omega$.} 
The flattened action chunk $\mathbf a_t\in\mathbb{R}^{d_{\mathbf a}}$ is embedded in two stages.
A pointwise linear projection, implemented as a $1\times1$ one-dimensional convolution, first maps $d_{\mathbf a}\!\rightarrow\!10$.
A two-layer MLP then maps $10\!\rightarrow\!768\!\rightarrow\!192$, with a SiLU activation between the two linear layers.
This produces an action embedding with the same width as the visual latent, $e_t^a=\psi_\omega(\mathbf a_t)\in\mathbb{R}^{192}.$

\paragraph{latent predictor $g_\phi$.} 
The predictor is a six-layer autoregressive Transformer with causal self-attention.
It operates at width $192$, uses $16$ attention heads with head dimension $64$, an MLP width of $2048$, attention and MLP dropout of $0.1$, and learned positional embeddings.
The predictor receives a sequence of latent tokens and a time-aligned sequence of action embeddings.
Actions condition every Transformer block through AdaLN-Zero\citep{peebles2023dit}.
In particular, each action embedding produces the shift, scale, and residual-gating parameters for both the attention and feed-forward sublayers; the modulation layer is initialized to zero.

The predictor uses at most the $m=3$ most recent latent-action pairs.
For a context ending at time $t$, its last output is interpreted as the prediction of the next latent,$\hat z_{t+1} =g_\phi\!\left( z_{t-m+1:t},\psi_\omega(\mathbf a_{t-m+1:t})\right),\label{eq:app-forward}$where the context is shortened near the beginning of a sequence.
The Transformer output is mapped to the latent space by a second projection MLP $192\!\rightarrow\!2048\!\rightarrow\!192$ with batch normalization.

\paragraph{Inverse-dynamics head $h_\xi$.} 
The inverse head receives the concatenation of a current latent and a future goal latent,
\begin{equation} h_\xi:\mathbb{R}^{2d}\rightarrow\mathbb{R}^{d_{\mathbf a}},\qquad\hat{\mathbf a}_i=h_\xi(z_i,z_{i+\kappa}).
\end{equation} It consists of three hidden layers of width $512$.
Each hidden layer applies a linear transformation, LayerNorm, and GELU activation.
A final linear layer maps to the model-level action dimension $d_{\mathbf a}$.
% For OGCube, we apply a $\tanh$ function to the output; for the other tasks, the output layer is linear.

\begin{table}[t]
    \centering
    \caption{Architecture used in all experiments.}
    \label{tab:app-architecture}
    \small
    \begin{tabular}{@{}lp{0.68\linewidth}@{}}
        \toprule
        Component & Configuration \\
        \midrule
        Latent width
        & $d=192$ \\
        Visual encoder $f_\theta$
        & ViT-Tiny; $12$ layers; $3$ heads; width $192$; patch size $14$;
          $224\times224$ inputs; trained from scratch \\
        Visual projection
        & MLP $192\!\rightarrow\!2048\!\rightarrow\!192$ with batch
          normalization \\
        Action encoder $\psi_\omega$
        & $1\times1$ convolution $d_{\mathbf a}\!\rightarrow\!10$, followed by
          MLP $10\!\rightarrow\!768\!\rightarrow\!192$ with SiLU \\
        latent predictor $g_\phi$
        & Causal Transformer; $6$ layers; $16$ heads; head dimension $64$;
          MLP width $2048$; dropout $0.1$; AdaLN-Zero conditioning \\
        Predictor context
        & $m=3$ latent--action pairs \\
        Predictor projection
        & MLP $192\!\rightarrow\!2048\!\rightarrow\!192$ with batch
          normalization \\
        Inverse head $h_\xi$
        & Three hidden layers of width $512$, each with LayerNorm and GELU
          % linear output to $d_{\mathbf a}$
          % ; task-dependent $\tanh$ 
          \\
        Frameskip
        & $k=5$ raw environment steps per model step \\
        \bottomrule
    \end{tabular}
\end{table}

\subsection{Training}
\label{app:training}
    
All four modules are optimized jointly.
Training combines next-latent prediction, anti-collapse representation regularization, hindsight-conditioned inverse dynamics, and action transition consistency.

\paragraph{latent prediction loss.}

Although every sampled window contains $W=16$ model steps, the prediction term uses a causal context of length $m=3$.
Specifically, the predictor consumes $z_{0:m-1}$ and the corresponding action embeddings and predicts the shifted latent sequence $z_{1:m}$.
Because causal attention is used, the output at position $t$ only depends on positions up to $t$.
Denoting the three predicted latents by $\hat z_{1:m}$, we compute
\begin{equation}
    \mathcal L_{\mathrm{pred}}
    =
    \frac{1}{Bmd}
    \sum_{b=1}^{B}\sum_{t=0}^{m-1}
    \left\|
        \hat z_{t+1}^{(b)}-z_{t+1}^{(b)}
    \right\|_2^2.
    \label{eq:app-pred-loss}
\end{equation}

The target latent in this term remains attached to the computational graph, for which we do not employ a momentum target encoder.

\paragraph{Anti-collapse regularization.}
We apply the Sketch Isotropic Gaussian Regularizer (SIGReg; \citealp{balestriero2025lejepa}) to the latents from all $W$ frames, including frames that are not used by Equation~\ref{eq:app-pred-loss}.
SIGReg measures the discrepancy between random one-dimensional projections of the latent distribution and a standard Gaussian reference distribution.
We use $17$ integration knots and $1024$ random projections.
Applying this objective over the complete window increases the number and temporal diversity of latent samples contributing to the regularizer.

\paragraph{Hindsight-conditioned inverse dynamics.} 
For each training window, we independently sample $N_a$ anchor--goal pairs.
For the $n$-th pair, the anchor is sampled uniformly from all positions with an available successor,
\begin{equation}
    i_n\sim\operatorname{Unif}\{0,\ldots,W-2\}.
\end{equation}

Conditioned on $i_n$, we sample a future offset uniformly from the remaining
part of the window,
\begin{equation}
    \kappa_n \sim
    \operatorname{Unif}\{1,\ldots,W-1-i_n\}.
    \label{eq:app-hindsight-sampling}
\end{equation}

The resulting goal index is $i_n+\kappa_n$.
Sampling is performed independently for each pair and each element of the minibatch, so anchors may repeat.
This procedure exposes the inverse head to goal conditions at multiple temporal distances while requiring no rewards or task labels.

For every sampled pair, the inverse head predicts the demonstrated action chunk executed at the anchor,
\begin{equation}
    \hat{\mathbf a}_{i_n}
    =h_\xi(z_{i_n},z_{i_n+\kappa_n}).
\end{equation}

Importantly, the target is the immediate action chunk $\mathbf a_{i_n}$, not the sequence of actions between the anchor and the sampled goal.
The inverse loss is the element-wise mean absolute error
\begin{equation}
    \mathcal L_{\mathrm{inv}}
    =
    \frac{1}{B N_a d_{\mathbf a}}
    \sum_{b=1}^{B}\sum_{n=1}^{N_a}
    \left\|
        \hat{\mathbf a}_{i_n}^{(b)}-
        \mathbf a_{i_n}^{(b)}
    \right\|_1.
    \label{eq:app-inv-loss}
\end{equation}

\paragraph{Action transition consistency loss.}

The action consistency term couples the inverse head to the learned latent prediction.
For each hindsight pair, we embed the predicted action and perform a single predictor step from the anchor latent,
\begin{equation}
    \tilde z_{i_n+1}
    =
    g_\phi\!\left(
        z_{i_n},
        \psi_\omega(\hat{\mathbf a}_{i_n})
    \right).
\end{equation}

We then regress the prediction toward the actual next latent using
\begin{equation}
    \mathcal L_{\mathrm{cyc}}
    =
    \frac{1}{B N_a d}
    \sum_{b=1}^{B}\sum_{n=1}^{N_a}
    \left\|
        \tilde z_{i_n+1}^{(b)}-
        \operatorname{sg}\!\left[z_{i_n+1}^{(b)}\right]
    \right\|_2^2,
    \label{eq:app-cycle-loss}
\end{equation}
where $\operatorname{sg}$ denotes stop-gradient.
Gradients from this loss update $h_\xi$, $\psi_\omega$, $g_\phi$, and $f_\theta$, as well as the anchor-side representation through $z_{i_n}$, while the next-latent target is detached.
The single-step context used by this term is supported by the same autoregressive predictor used for the length-$m$ forward objective.

\paragraph{Joint objective.}
The complete objective is
\begin{equation}
    \mathcal L
    =
    \mathcal L_{\mathrm{pred}}
    +\lambda_{\mathrm{reg}}\mathcal L_{\mathrm{sigreg}}
    +\lambda_{\mathrm{inv}}\mathcal L_{\mathrm{inv}}
    +\lambda_{\mathrm{cyc}}\mathcal L_{\mathrm{cyc}}.
    \label{eq:app-total-loss}
\end{equation}
All terms are averaged over the batch and their corresponding temporal, anchor, action, or latent dimensions before the weights are applied.

\paragraph{Optimization.}
 We use a single AdamW optimizer for $f_\theta$, $\psi_\omega$, $g_\phi$, and $h_\xi$, with weight decay $10^{-3}$.
Training uses bf16 mixed precision and a global gradient-norm clipping threshold of $1.0$.
The learning rate follows a linear-warmup cosine-decay schedule, starting from zero.
With ten training epochs, the configured warmup is rounded to one epoch, followed by cosine decay to zero.
Training and validation use the LeWM-based window-level split
described in the dataset appendix \ref{app:datasets}, with a 90\%/10\% ratio and
random seed $3072$.

Algorithm~\ref{alg:training} summarizes the complete training procedure, and Table~\ref{tab:training_hyperparameters} reports the task-specific training parameters.

\begin{table*}[t]
\centering
\small
\caption{
Task-specific training hyperparameters for HIJP.}
\label{tab:training_hyperparameters}
\begin{tabular}{lcccccccc}
\toprule
Task
& Batch
& Learning rate
& $N_a$
& $\lambda_{\mathrm{reg}}$
& $\lambda_{\mathrm{inv}}$
& $\lambda_{\mathrm{cyc}}$
& Epochs
& Scheduler \\
\midrule
TwoRoom & 48 & $3\times10^{-5}$ & 4 & 0.09 & 1.0 & 0.05 & 10 & LWC \\
Reacher & 48 & $5\times10^{-5}$ & 4 & 0.09 & 1.0 & 0.10 & 10 & LWC \\
PushT   & 48 & $5\times10^{-5}$ & 4 & 0.09 & 1.0 & 0.10 & 10 & LWC \\
OGCube  & 32 & $5\times10^{-5}$ & 4 & 0.09 & 1.0 & 0.10 & 10 & LWC \\
\bottomrule
\end{tabular}
\end{table*}

\subsection{Planning and Receding-Horizon Execution}
\label{app:planning}
At each replanning point, the current observation and target observation are encoded as
\begin{equation}
    z_0=f_\theta(o_t),
    \qquad
    z_g=f_\theta(o_g).
\end{equation}

Starting from $z_0$, we construct an $H$-step action sequence through the autoregressive latent rollout introduced in the main text.
At rollout step $j$, the inverse head first predicts the action chunk that is consistent with moving from the current imagined latent toward the goal latent,
\begin{equation}
    \mathbf a_j=h_\xi(z_j,z_g).
\end{equation}
The action chunk is embedded and passed to the latent predictor, which advances the imagined state,
\begin{equation}
    z_{j+1}
    =g_\phi\!\left(
        z_{\max(0,j-m+1):j},
        \psi_\omega(\mathbf a_{\max(0,j-m+1):j})
    \right).
    \label{eq:app-planning-rollout}
\end{equation}

The available context grows from one to $m$ latent-action pairs during the first rollout steps and is subsequently truncated to the most recent $m=3$ pairs.
Thus, constructing an $H$-step plan requires $H$ evaluations of the inverse head and $H$ autoregressive evaluations of the latent predictor.
No population of candidate action sequences or iterative test-time action optimization is required.

We use $H=5$ model steps for every task.
Since each model-level action contains $k=5$ atomic actions, one predicted plan covers $Hk=25$ raw environment steps.
The controller operates in a receding-horizon loop with execution stride $R=5$ model steps.
It therefore executes the complete predicted action sequence, re-encodes the resulting observation, and constructs a new plan.
The target observation remains unchanged within an episode but is encoded together with the current observation at each replanning point.
Execution terminates when the task-specific success condition is satisfied, or the allotted budget of raw environment steps is exhausted. Algorithm~\ref{alg:planning} gives the complete procedure. 

Table~\ref{tab:app-planning} collects the planning settings shared across the
four tasks. Keeping these settings unchanged ensures that differences across
tasks do not arise from task-specific planning budgets or model-rollout depths.

\begin{table}[htbp]
    \centering
    \caption{Planning settings shared across all four tasks.}
    \label{tab:app-planning}
    \small
    \begin{tabular}{@{}lc@{}}
        \toprule
        Hyperparameter & Value \\
        \midrule
        Planning horizon $H$ & $5$ model steps \\
        Action-chunk length $k$ & $5$ raw steps \\
        Raw planning horizon $Hk$ & $25$ environment steps \\
        Execution stride $R$ & $5$ model steps \\
        Predictor context $m$ & $3$ latent--action pairs \\
        \bottomrule
    \end{tabular}
\end{table}

% ---------- Algorithm 1: Training ----------
\begin{algorithm}[t]
\caption{Joint training with hindsight-conditioned inverse dynamics}
\label{alg:training}
\begin{algorithmic}[1]
\Require Offline dataset $\mathcal D$; window length $W$;
predictor context length $m$; number of hindsight pairs $N_a$

\For{each training minibatch}
    \State Sample a minibatch of windows
    $(o_{0:W-1},\mathbf a_{0:W-2})\sim\mathcal D$

    \State Encode observations and actions:
    $z_t\gets f_\theta(o_t)$ and
    $e_t^a\gets\psi_\omega(\mathbf a_t)$

    \State Predict the next latents:
    $\hat z_{1:m}\gets
    g_\phi(z_{0:m-1},e^a_{0:m-1})$

    \State $\mathcal L_{\mathrm{pred}}
    \gets\operatorname{MSE}(\hat z_{1:m},z_{1:m})$

    \State $\mathcal L_{\mathrm{sigreg}}
    \gets\operatorname{SIGReg}(z_{0:W-1})$

    \For{$n=1,\ldots,N_a$}
        \State Sample
        $i_n\sim\operatorname{Unif}\{0,\ldots,W-2\}$

        \State Sample
        $\kappa_n\sim
        \operatorname{Unif}\{1,\ldots,W-1-i_n\}$

        \State Predict the demonstrated action:
        $\hat{\mathbf a}_{i_n}
        \gets h_\xi(z_{i_n},z_{i_n+\kappa_n})$

        \State Predict the successor latent:
        $\tilde z_{i_n+1}
        \gets
        g_\phi\!\left(
        z_{i_n},
        \psi_\omega(\hat{\mathbf a}_{i_n})
        \right)$
    \EndFor

    \State $\mathcal L_{\mathrm{inv}}
    \gets
    \operatorname{MAE}\!\left(
    \{\hat{\mathbf a}_{i_n}\}_{n=1}^{N_a},
    \{\mathbf a_{i_n}\}_{n=1}^{N_a}
    \right)$

    \State $\mathcal L_{\mathrm{cyc}}
    \gets
    \operatorname{MSE}\!\left(
    \{\tilde z_{i_n+1}\}_{n=1}^{N_a},
    \{\operatorname{sg}[z_{i_n+1}]\}_{n=1}^{N_a}
    \right)$

    \State $\mathcal L\gets
    \mathcal L_{\mathrm{pred}}
    +\lambda_{\mathrm{reg}}\mathcal L_{\mathrm{sigreg}}
    +\lambda_{\mathrm{inv}}\mathcal L_{\mathrm{inv}}
    +\lambda_{\mathrm{cyc}}\mathcal L_{\mathrm{cyc}}$

    \State Update $(\theta,\omega,\phi,\xi)$ with AdamW
\EndFor
\end{algorithmic}
\end{algorithm}

% ---------- Algorithm 2: Planning ----------
\begin{algorithm}[t]
\caption{Amortized latent planning}
\label{alg:planning}
\begin{algorithmic}[1]
\Require Trained $f_\theta$, $\psi_\omega$, $g_\phi$, and $h_\xi$;
goal observation $o_g$; planning horizon $H$;
execution stride $R$; action-chunk length $k$;
predictor context length $m$; raw-step budget $B$

\State $n\gets0$

\While{$n<B$ and the task has not succeeded}
    \State Encode the current and goal observations:
    $\hat z_0\gets f_\theta(o_t)$,
    $z_g\gets f_\theta(o_g)$

    \State Initialize
    $\mathcal Z\gets[\hat z_0]$ and
    $\mathcal A\gets[\,]$

    \For{$j=0,\ldots,H-1$}
        \State Predict an action chunk:
        $\mathbf a_j\gets h_\xi(\hat z_j,z_g)$

        \State Append $\mathbf a_j$ to $\mathcal A$

        \State Predict the next imagined latent:
        \[
        \hat z_{j+1}\gets
        g_\phi\!\left(
        \operatorname{tail}_m(\mathcal Z),
        \psi_\omega\!\left(
        \operatorname{tail}_m(\mathcal A)
        \right)
        \right)
        \]

        \State Append $\hat z_{j+1}$ to $\mathcal Z$
    \EndFor

    \State
    $q\gets
    \min\left\{
    R,\,
    H,\,
    \left\lfloor\frac{B-n}{k}\right\rfloor
    \right\}$

    \State Execute the first $q$ action chunks in $\mathcal A$,
    terminating early if the task succeeds

    \State Let $s$ be the number of raw environment steps
    actually executed

    \State $n\gets n+s$
\EndWhile
\end{algorithmic}
\end{algorithm}

\section{Loss Weight Ablation}
\label{app:loss_weight_ablation}

To examine the sensitivity of HIJP to the weighting coefficients of its training objectives, we conduct ablations on the inverse dynamics loss weight $\lambda_{\mathrm{inv}}$ and the cycle-consistency loss weight $\lambda_{\mathrm{cyc}}$.

Specifically, we vary $\lambda_{\mathrm{inv}} \in \{0, 0.1, 1.0, 2.0\}$ while fixing $\lambda_{\mathrm{cyc}}=0.1$, and vary $\lambda_{\mathrm{cyc}} \in \{0, 0.01, 0.1, 0.2\}$ while fixing $\lambda_{\mathrm{inv}}=1.0$.
We train a separate model for each weight setting and evaluate the resulting checkpoint using the same evaluation protocol as in the main experiments. All training and evaluation settings follow Section \ref{sec:exp-setup}.
We report the results on Reacher, PushT, and OGCube in Table \ref{tab:loss_weight_ablation}.
We omit TwoRoom because its performance is already saturated under our evaluation protocol, with near 100\% success rates, making it uninformative for distinguishing different loss-weight choices.

\begin{table}[t]
\centering
\small
\caption{Success rates (\%) for the loss-weight ablations. Avg. reports the arithmetic mean across the four difficulty tiers, and Overall Avg. averages all 12 task--tier combinations. Bold column headers indicate the default weights. In the Overall Avg. row, the best and second-best results within each weight group are shown in \textbf{bold} and \underline{underlined text}.}
\label{tab:loss_weight_ablation}
\setlength{\tabcolsep}{3.5pt}
\fontsize{7.75pt}{8.225pt}\selectfont
\renewcommand{\arraystretch}{0.96}

\begin{tabular}{@{}ll cccc cccc@{}}
\toprule
& &
\multicolumn{4}{c}{
$\lambda_{\mathrm{inv}}$ ($\lambda_{\mathrm{cyc}}=0.1$)}
&
\multicolumn{4}{c}{
$\lambda_{\mathrm{cyc}}$ ($\lambda_{\mathrm{inv}}=1.0$)}
\\
\cmidrule(lr){3-6}
\cmidrule(lr){7-10}
Task
& Tier
& 0
& 0.1
& \textbf{1.0}
& 2.0
& 0
& 0.01
& \textbf{0.1}
& 0.2
\\
\midrule

% ======================== Reacher ========================
\multirow{5}{*}{Reacher}
& Easy
& 36\%
& 42\%
& 48\%
& 44\%
& 34\%
& 52\%
& 48\%
& 44\%
\\
& Default
& 72\%
& 88\%
& 92\%
& 78\%
& 78\%
& 86\%
& 92\%
& 64\%
\\
& Hard
& 90\%
& 86\%
& 92\%
& 90\%
& 84\%
& 90\%
& 92\%
& 78\%
\\
& Extreme
& 86\%
& 98\%
& 94\%
& 90\%
& 94\%
& 90\%
& 94\%
& 92\%
\\
\cmidrule(l){2-10}
\rowcolor{avgrow}
& Avg.
& 71.0\%
& 78.5\%
& 81.5\%
& 75.5\%
& 72.5\%
& 79.5\%
& 81.5\%
& 69.5\%
\\
\midrule

% ======================== PushT ========================
\multirow{5}{*}{PushT}
& Easy
& 36\%
& 90\%
& 90\%
& 92\%
& 84\%
& 90\%
& 90\%
& 92\%
\\
& Default
& 6\%
& 78\%
& 80\%
& 82\%
& 78\%
& 72\%
& 80\%
& 82\%
\\
& Hard
& 0\%
& 56\%
& 66\%
& 62\%
& 54\%
& 56\%
& 66\%
& 68\%
\\
& Extreme
& 0\%
& 36\%
& 42\%
& 58\%
& 22\%
& 58\%
& 42\%
& 44\%
\\
\cmidrule(l){2-10}
\rowcolor{avgrow}
& Avg.
& 10.5\%
& 65.0\%
& 69.5\%
& 73.5\%
& 59.5\%
& 69.0\%
& 69.5\%
& 71.5\%
\\
\midrule

% ======================== OGCube ========================
\multirow{5}{*}{OGCube}
& Easy
& 80\%
& 94\%
& 98\%
& 100\%
& 98\%
& 96\%
& 98\%
& 92\%
\\
& Default
& 56\%
& 94\%
& 96\%
& 94\%
& 96\%
& 94\%
& 96\%
& 92\%
\\
& Hard
& 46\%
& 90\%
& 96\%
& 96\%
& 94\%
& 96\%
& 96\%
& 94\%
\\
& Extreme
& 40\%
& 96\%
& 96\%
& 96\%
& 96\%
& 94\%
& 96\%
& 94\%
\\
\cmidrule(l){2-10}
\rowcolor{avgrow}
& Avg.
& 55.5\%
& 93.5\%
& 96.5\%
& 96.5\%
& 96.0\%
& 95.0\%
& 96.5\%
& 93.0\%
\\

\specialrule{0.08em}{0.35em}{0.25em}

% ======================== Overall ========================
\rowcolor{avgrow}
\multicolumn{2}{@{}l}{\textbf{Overall Avg.}}
& 45.7\%
& 79.0\%
& \best{82.5\%}
& \snd{81.8\%}
& 76.0\%
& \snd{81.2\%}
& \best{82.5\%}
& 78.0\%
\\
\bottomrule
\end{tabular}
\end{table}
    
The default setting, $\lambda_{\mathrm{inv}}=1.0$ and $\lambda_{\mathrm{cyc}}=0.1$, achieves the highest overall average success rate of 82.5\%.
Increasing $\lambda_{\mathrm{inv}}$ to 2.0 improves performance on PushT and matches the default performance on OGCube, but reduces performance on Reacher, resulting in a slightly lower overall average of 81.8\%. 
Similarly, decreasing $\lambda_{\mathrm{cyc}}$ to 0.01 or increasing it to 0.2 yields overall averages of 81.2\% and 78.0\%, respectively.
These results indicate that the default weights provide the best overall balance across the three evaluated environments.

% \section{Inverse-Dynamics Diagnostic Probes}
% \label{app:mechanism}
% \input{text/app_inverse_dynamics}

\section{Action Smoothness Experiment}
\label{app:smoothness}
This appendix provides the complete evaluation protocol, metric definition, and per-tier results for the action-smoothness study in Section~\ref{sec:exp-smoothness}. We evaluate HIJP, its CEM ablation, and LeWM on the four domains described in Appendix~\ref{app:datasets}.

For each domain and difficulty tier, all methods are evaluated closed-loop on the same set of start--goal instances. Instance sampling depends only on the dataset, random seed, goal offset, and number of evaluation episodes, and is therefore independent of the evaluated method. 
We use 50 episodes per task with random seed 42. The four difficulty tiers are defined by the goal-offset and step-budget pairs $(K, B)$ listed in Table~\ref{tab:difficulty}. 
Each rollout is allowed to run for the corresponding step budget, and we record the action executed at every environment step.

HIJP directly executes the amortized plan produced by its inverse-guided controller and performs no test-time search. 
The CEM ablation uses the same learned representation and predictive model as HIJP, but replaces inverse-guided action generation with CEM planning. 
LeWM, which performs CEM planning over its own forward model, is included as an additional reference.
All methods use the same control frequency and normalized action space.

% 需要: \usepackage{booktabs}  \usepackage[table]{xcolor}
\begin{table}[h]
\centering
\caption{ Action jerk across four environments and four difficulty tiers. Avg. reports the arithmetic mean across tiers, with average ratios computed from the per-tier ratios. The lowest jerk in each row is shown in \textbf{bold}.
}
\label{tab:jerk-appendix}
\begin{tabular}{llccccc}
\toprule
\fontsize{7.75pt}{8.225pt}\selectfont
& & \textbf{HIJP} & \multicolumn{2}{c}{CEM (ablation)} & \multicolumn{2}{c}{LeWM} \\
\cmidrule(lr){3-3}\cmidrule(lr){4-5}\cmidrule(lr){6-7}
Environment & Tier & Jerk $\downarrow$ & Jerk $\downarrow$ & Ratio $\uparrow$ & Jerk $\downarrow$ & Ratio $\uparrow$ \\
\midrule
\multirow{5}{*}{PushT}
 & easy    & \textbf{0.038} & 0.439 & 11.5$\times$ & 0.451 & 11.8$\times$ \\
 & default & \textbf{0.040} & 0.450 & 11.2$\times$ & 0.407 & 10.1$\times$ \\
 & hard    & \textbf{0.048} & 0.611 & 12.8$\times$ & 0.595 & 12.4$\times$ \\
 & extreme & \textbf{0.048} & 0.697 & 14.7$\times$ & 0.724 & 15.2$\times$ \\
 \rowcolor{gray!12}
 & Avg.    & \textbf{0.043} & 0.549 & 12.6$\times$ & 0.544 & 12.4$\times$ \\
\midrule
\multirow{5}{*}{TwoRoom}
 & easy    & \textbf{0.082} & 1.838 & 22.4$\times$ & 1.774 & 21.6$\times$ \\
 & default & \textbf{0.084} & 2.216 & 26.3$\times$ & 2.191 & 26.0$\times$ \\
 & hard    & \textbf{0.120} & 2.325 & 19.4$\times$ & 2.428 & 20.2$\times$ \\
 & extreme & \textbf{0.146} & 2.503 & 17.1$\times$ & 2.558 & 17.5$\times$ \\
 \rowcolor{gray!12}
 & Avg.    & \textbf{0.108} & 2.220 & 21.3$\times$ & 2.238 & 21.3$\times$ \\
\midrule
\multirow{5}{*}{Reacher}
 & easy    & \textbf{0.040} & 1.179 & 29.5$\times$ & 1.099 & 27.5$\times$ \\
 & default & \textbf{0.046} & 1.174 & 25.6$\times$ & 1.114 & 24.3$\times$ \\
 & hard    & \textbf{0.046} & 1.200 & 26.3$\times$ & 1.134 & 24.8$\times$ \\
 & extreme & \textbf{0.046} & 1.222 & 26.4$\times$ & 1.134 & 24.5$\times$ \\
 \rowcolor{gray!12}
 & Avg.    & \textbf{0.045} & 1.194 & 27.0$\times$ & 1.120 & 25.3$\times$ \\
\midrule
\multirow{5}{*}{OGCube}
 & easy    & \textbf{0.126} & 1.596 & 12.7$\times$ & 1.517 & 12.1$\times$ \\
 & default & \textbf{0.073} & 1.843 & 25.2$\times$ & 1.886 & 25.8$\times$ \\
 & hard    & \textbf{0.088} & 1.848 & 21.0$\times$ & 2.159 & 24.5$\times$ \\
 & extreme & \textbf{0.099} & 2.065 & 20.9$\times$ & 2.304 & 23.3$\times$ \\
 \rowcolor{gray!12}
 & Avg.    & \textbf{0.097} & 1.838 & 19.9$\times$ & 1.967 & 21.4$\times$ \\
\bottomrule
\end{tabular}
\end{table}

If an episode reaches its goal before exhausting the step budget, we truncate its action sequence at the first success step and compute smoothness only over the executed prefix. 
This avoids either rewarding or penalizing post-success idling and measures the controls actually required to complete the task.
Unsuccessful episodes contribute their full budget-length action sequences.

Let $a_{1:T}=(a_1,\ldots,a_T)$ denote the executed action sequence of an episode. 
We define \emph{action jerk} as the mean $\ell_2$ norm of the discrete second difference:

\begin{equation}
    \operatorname{Jerk}(a_{1:T})
    =
    \frac{1}{T-2}
    \sum_{t=3}^{T}
    \left\lVert
        a_t - 2a_{t-1} + a_{t-2}
    \right\rVert_2 .
    \label{eq:jerk}
\end{equation}  

Lower values indicate smoother controls.
A temporally constant or linearly varying action sequence has zero second difference, whereas oscillatory or abruptly changing commands produce a large value.

For each method, environment, and difficulty tier, we compute
Eq. ~\ref{eq:jerk} independently for every episode and report the mean across episodes. Let $\overline{J}_{m}$ denote this episode-averaged jerk for method $m$. 
We additionally report
\begin{equation}
    R_{\mathrm{CEM}}
    =
    \frac{\overline{J}_{\mathrm{CEM}}}
         {\overline{J}_{\mathrm{HIJP}}},
    \qquad
    R_{\mathrm{LeWM}}
    =
    \frac{\overline{J}_{\mathrm{LeWM}}}
         {\overline{J}_{\mathrm{HIJP}}}.
    \label{eq:jerk-ratios}
\end{equation}

A ratio greater than one indicates that HIJP produces smoother actions than the corresponding CEM-based planner. Because all methods are evaluated in the same normalized action space, their jerk values are directly comparable.

Table~\ref{tab:jerk-appendix} reports the results across all four difficulty tiers. HIJP consistently achieves lower action jerk than both its CEM ablation and LeWM. 
On the extreme tier ($K=75$), the CEM ablation produces $14.7\times$--$26.4\times$ higher jerk than HIJP, despite sharing the same learned representation and predictive model. 
This controlled comparison shows that the smoothness advantage primarily arises from HIJP's amortized inverse-guided planning rather than from a difference in learned representations. 
LeWM exhibits a similar level of jerk to the CEM ablation, suggesting that independently optimizing successive plans with CEM can introduce substantial temporal discontinuities into the executed controls.

\section{Visualizing Planned Trajectories}
\label{app:qualitative}

\begin{figure}[htbp]
\centering
\includegraphics[width=\linewidth]{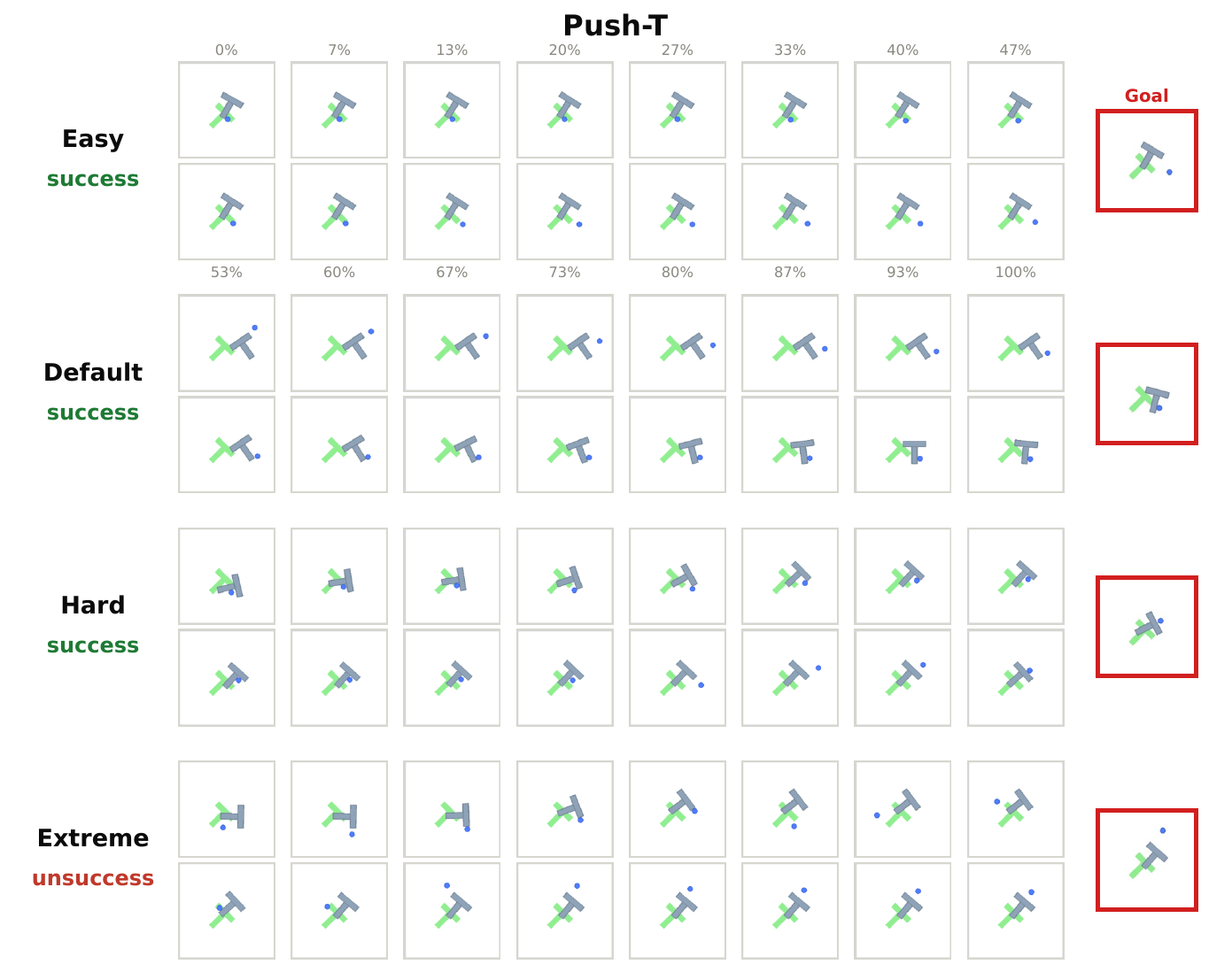}
\caption{Representative trajectories produced by HIJP in the Push-T environment.}
\label{fig:qual-pusht}
\end{figure}

\begin{figure}[htbp]
\centering
\includegraphics[width=\linewidth]{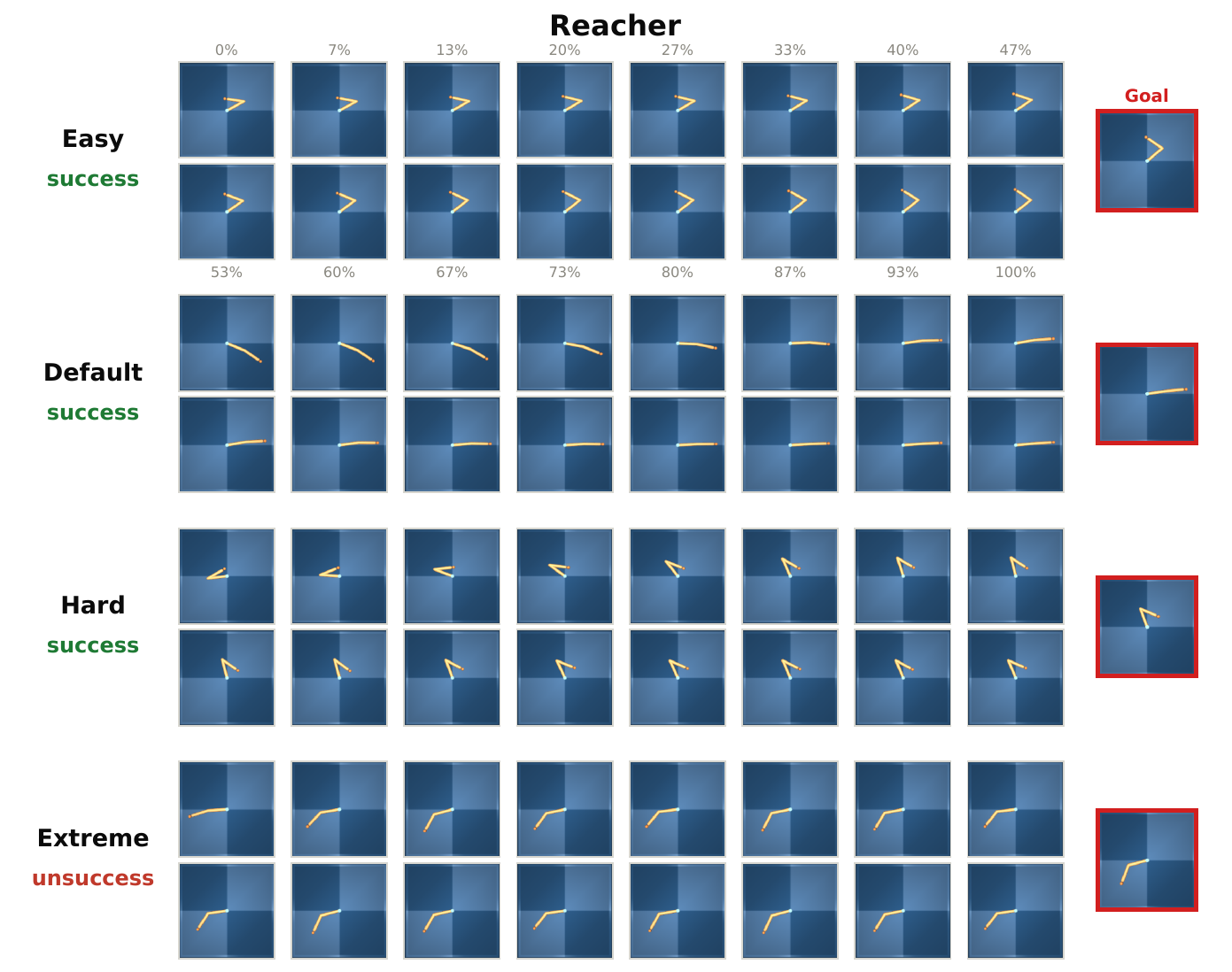}
\caption{Representative trajectories produced by HIJP in the Reacher environment.}
\label{fig:qual-reacher}
\end{figure}

\begin{figure}[htbp]
\centering
\includegraphics[width=\linewidth]{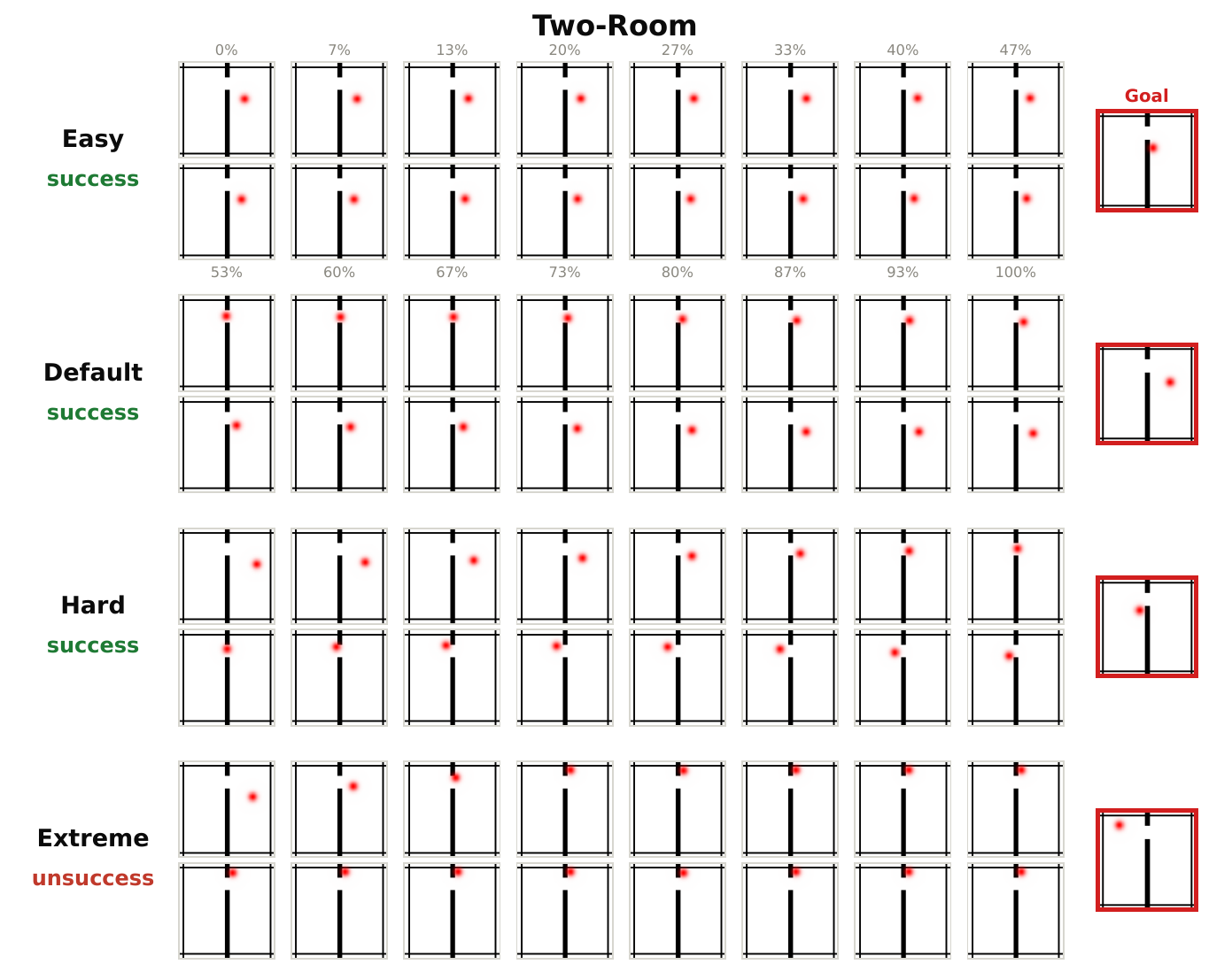}
\caption{Representative trajectories produced by HIJP in the TwoRoom environment.}
\label{fig:qual-tworoom}
\end{figure}

\begin{figure}[htbp]
\centering
\includegraphics[width=\linewidth]{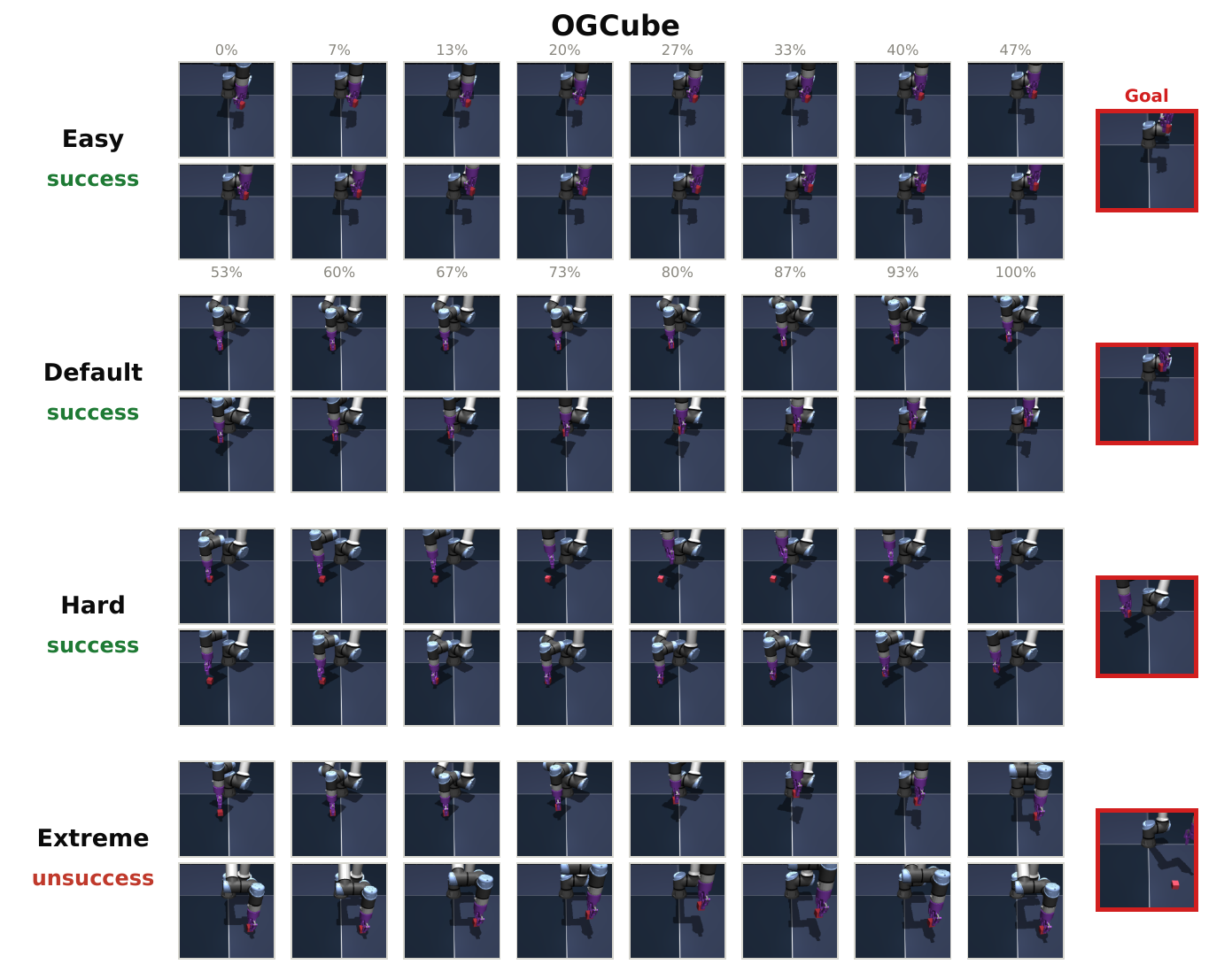}
\caption{Representative trajectories produced by HIJP in the OGCube environment.}
\label{fig:qual-cube}
\end{figure}

For each environment, we run our planner (Algorithm~\ref{alg:planning}) from a start observation toward a goal observation and record the resulting trajectory.
Figures~\ref{fig:qual-pusht}--\ref{fig:qual-cube} visualize these
rollouts as pixel-space filmstrips.

Each figure corresponds to one environment and contains four difficulty tiers: Easy, Default, Hard, and Extreme.
The tiers are defined by goal offsets of $10$, $25$, $50$, and $75$ environment steps, with corresponding planning budgets of $25$, $50$, $100$, and $100$ steps, respectively.
For each tier, we show one rollout using $16$ frames sampled uniformly over its execution, from $0\%$ to $100\%$, and arrange them in two rows of eight frames.
The goal observation appears separately on the right, outlined in red.
The label beside each tier indicates whether the rollout reaches the goal within the planning budget.

\end{document}